\documentclass[table]{article} 
\usepackage{iclr2024_conference,times}

\usepackage{amsmath,amsfonts,bm}

\def\eqref#1{equation~\ref{#1}}

\def\1{\bm{1}}

\DeclareMathAlphabet{\mathsfit}{\encodingdefault}{\sfdefault}{m}{sl}
\SetMathAlphabet{\mathsfit}{bold}{\encodingdefault}{\sfdefault}{bx}{n}

\newcommand{\KL}{D_{\mathrm{KL}}}

\usepackage[utf8]{inputenc} 
\usepackage[T1]{fontenc}    
\usepackage{hyperref}       
\usepackage{url}            
\usepackage{booktabs}       
\usepackage{amsfonts}       
\usepackage{nicefrac}       
\usepackage{microtype}      
\usepackage{highlight}

\usepackage{cleveref}
\usepackage{standalone}
\usepackage{latexsym}
\usepackage{amsmath}
\usepackage{amssymb}
\usepackage{amsthm}
\usepackage{graphicx}
\usepackage{subcaption}
\usepackage{array}
\usepackage{tabu}
\usepackage{makecell}
\usepackage{paralist}
\usepackage{cases}
\usepackage{diagbox}
\usepackage{enumitem}
\usepackage{soul}
\usepackage{multirow}
\usepackage{verbatim}
\usepackage{tabulary}
\usepackage{booktabs}
\usepackage[mathscr]{euscript}
\usepackage{mathtools}
\usepackage{algorithm}
\usepackage{algorithmic}
\usepackage{stmaryrd}
\usepackage{tikz-dependency}
\usetikzlibrary{automata,decorations.markings,arrows,positioning,matrix,calc,patterns,angles,quotes,calc}
\usepackage{adjustbox}
\usepackage{tabularx}
\usepackage{xspace}
\usepackage{tabulary}
\usepackage{afterpage}
\usepackage{bm}
\usepackage{color}
\usepackage{graphicx}
\usepackage{slashbox}
\usepackage[toc,page]{appendix}
\usepackage{makecell}
\usepackage{boldline}
\usepackage[flushleft]{threeparttable}
\usepackage[shortcuts]{extdash}  
\usepackage{blindtext}
\usepackage{graphicx}
\usepackage{capt-of}
\usepackage{booktabs}
\usepackage{varwidth}
\usepackage{pifont}
\usepackage{wrapfig}
\usepackage{listings}

\usepackage{pythonhighlight}

\definecolor{orange}{rgb}{1,0.5,0}
\definecolor{mdgreen}{rgb}{0.05,0.6,0.05}
\definecolor{mdblue}{rgb}{0,0,0.7}
\definecolor{dkblue}{rgb}{0,0,0.5}
\definecolor{dkgray}{rgb}{0.3,0.3,0.3}
\definecolor{slate}{rgb}{0.25,0.25,0.4}
\definecolor{gray}{rgb}{0.5,0.5,0.5}
\definecolor{ltgray}{rgb}{0.7,0.7,0.7}
\definecolor{purple}{rgb}{0.7,0,1.0}
\definecolor{lavender}{rgb}{0.65,0.55,1.0}

\definecolor{mypurple}{RGB}{111,61,121}
\definecolor{myblue}{RGB}{46,88,180}
\definecolor{myred}{RGB}{181,68,106}
\definecolor{myyellow}{RGB}{204,143,55}

\newcommand{\paragraphmini}[1]{\noindent\textbf{#1}}
\creflabelformat{equation}{#2#1#3}

\newcommand{\interalia}[1]{\citep[\emph{inter alia}]{#1}}

\DeclareSymbolFont{extraup}{U}{zavm}{m}{n}
\DeclareMathSymbol{\vardiamond}{\mathalpha}{extraup}{87}

\newcolumntype{L}[1]{>{\raggedright\let\newline\\\arraybackslash\hspace{0pt}}m{#1}}
\newcolumntype{C}[1]{>{\centering\let\newline\\\arraybackslash\hspace{0pt}}m{#1}}
\newcolumntype{R}[1]{>{\raggedleft\let\newline\\\arraybackslash\hspace{0pt}}m{#1}}

\theoremstyle{definition}

\theoremstyle{remark}

\setul{1pt}{.4pt}

\DeclareFixedFont{\ttb}{T1}{txtt}{bx}{n}{12} 
\DeclareFixedFont{\ttm}{T1}{txtt}{m}{n}{12}  

\definecolor{applegreen}{rgb}{0.55, 0.71, 0.0}

\title{TRAM:\@ Bridging Trust Regions and Sharpness Aware Minimization} 

\author{ {Tom Sherborne$^{1}$\thanks{\hspace{1mm}This work was done while Tom
    Sherborne and Hao Peng were at the Allen Institute for AI.}\quad Naomi
    Saphra$^{2}$ \quad Pradeep Dasigi$^{3}$ \quad Hao Peng$^{4*}$} \\
$^1$University of Edinburgh \quad $^2$Kempner Institute, Harvard University
\quad $^3$Allen Institute for AI  \\
$^4$University of Illinois Urbana-Champaign \\
\texttt{tom.sherborne@ed.ac.uk,~nsaphra@fas.harvard.edu} \\
\texttt{pradeepd@allenai.org,~haopeng@illinois.edu} 
}

\iclrfinalcopy
\begin{document}

\maketitle

\begin{abstract}
Sharpness-aware minimization (SAM) reports improving domain generalization by
reducing the loss surface curvature in the parameter space. However,
generalization during \textit{fine-tuning} is often more dependent on the
transferability of \textit{representations} in the function space. Trust-region
methods (TR) target this goal by regularizing representation curvature to reduce
catastrophic forgetting of pre-trained task-agnostic information while adopting
task-specific skills. We consider unifying these strategies for low curvature in
both parameter space and function space to improve out-of-domain (OOD)
generalization. We propose \textbf{Trust Region Aware Minimization} (TRAM), a
SAM algorithm fine-tuning for low parameter sharpness and smooth, informative
representations preserving pre-trained structure. TRAM uses a trust region bound
to inform the SAM adversarial neighborhood, introducing an awareness of function
curvature within optimization for flatter minima. We empirically validate TRAM
in vision (cross-dataset adaptation) and text (OOD language modeling, zero-shot
cross-lingual transfer) tasks where robust domain transfer and representation
generality are critical. TRAM outperforms SAM- and TR-based optimization across
all tasks, notably surpassing competing methods for hard transfer between
\textit{anticorrelated} domains. TRAM establishes a novel standard in
fine-tuning for domain-generalizable models with minimal additional computation
over previous sharpness-aware methods. 


\end{abstract}

\section{Introduction}

Neural model training requires navigating over a complex, non-convex loss
surface \citep{Frankle2020RevisitingC} towards a good local minimum. Studying
loss surfaces and training dynamics has led to many algorithmic advances
\citep{izmailov2018averaging,foret2021sharpnessaware-SAM,lion} and
regularization schemes \citep{JMLR:v15:srivastava14a-dropout,
pmlr-v37-ioffe15-bn} to improve optimization. One such strategy is to exploit an
association between generalization and flat minima, defined by
\citet{NIPS1994_01882513-flat-minima} as ``region[s] in weight space with the
property that each weight vector from that region has [a] similar small error''.
Intuitively, a flatter, or less sharp~\citep{keskar2017on}, minimum will
generalize better, as the loss function will be non-increasing under
distribution shift. Recent work has developed a family of \emph{sharpness-aware
minimization} (SAM) algorithms targeting flat minima by jointly minimizing a
worst-case generalization bound and local parameter sharpness
\citep{foret2021sharpnessaware-SAM,pmlr-v139-kwon21b-ASAM,pmlr-v162-kim22f-FisherSAM,mollenhoff2023bayesian-sam}. 
 
While flat minima methods report widespread improvement over conventional
optimizers \citep{kaddour-2022-when-do-flat-minima-opt}, we argue that they are
not fully connected to the modern \emph{fine-tuning} paradigm, wherein a
task-specific model inherits parameters from a pre-trained model instead of
being trained from scratch \citep{wang2018glue,liang-etal-2020-xglue}. In these
settings, focusing on local properties of the loss landscape (e.g., sharpness)
may fail by suboptimally exploiting useful generic task-agnostic structures
within pre-trained representations. In this work, we propose to combine
sharpness-aware minimization with the robust transfer of pre-trained information
(in representation space) for fine-tuning scenarios requiring
out-of-distribution knowledge for successful adaptation.

\begin{wrapfigure}{L}{0.5\textwidth}
    \centering
    \vspace{-1em}
    \includegraphics[width=0.5\textwidth]{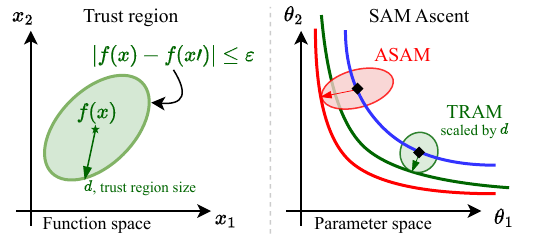}
    \caption{TRAM introduces an awareness of function curvature (i.e., the trust
    region) into sharpness-aware minimization. (left) TRAM estimates the size of the
    trust region, $d$, around $f\left(x\right)$ in {\color{applegreen} green}.
    (right) the loss contour in parameter space following \citet{pmlr-v139-kwon21b-ASAM}
    where {\color{blue}blue} is the typical loss; {\color{red}{red}} is the maximized 
    worst-case loss for ASAM; and {\color{applegreen}{green}} is the maximized loss
    within the subdomain constrained for function smoothness.
    }\label{fig:example}
    \vspace{-1em}
\end{wrapfigure}

\vspace{-1em}
Existing methods to improve leveraging pre-trained structure during fine-tuning
include trust region regularization
\citep{pmlr-v37-schulman15-trust-region-policy,jiang-etal-2020-smart,
aghajanyan2021better} or adversarial perturbation
\citep{Zhu2020FreeLB,he2021deberta}. These methods focus on the curvature of the
function itself e.g., by encouraging smooth local changes in representations.
The intuition is that lower representation curvature during fine-tuning limits a
function from catastrophically forgetting \interalia{FRENCH1999128} useful
information from pre-training. This representation smoothing approach contrasts
with SAM-style optimization for parameter smoothness. Both perspectives show
empirical improvement in downstream tasks
\citep{aghajanyan2021better,bahri-etal-2022-sharpness}, but a fusion of these
strategies is presently under-explored.

To this end, we propose \textbf{TRAM}:~Trust Region Aware Minimization, a
fine-tuning algorithm for out-of-distribution generalization combining the
success of both sharpness-aware and trust region optimization. TRAM uses a trust
region bound to inform the SAM adversarial neighborhood, introducing an
awareness of function curvature within optimization for flatter minima. The
resulting algorithm yields low-sharpness parameters and improved adaptation of
pre-trained models to downstream tasks. To illustrate TRAM's advantage over
strong baselines in retaining generic representations, we focus on
\emph{distribution transfer} challenges within Transformer-based models. Our
contributions are:

\begin{itemize}
    \item We propose a new optimization algorithm: \textbf{Trust Region Aware
    Minimization} integrates representation smoothing regularization into
    sharpness-aware minimization. We propose and contrast multiple variants of
    TRAM based on differing perspectives on trust region estimation and
    efficiency trade-offs (\Cref{sec:tram}).\footnote{Code at
    \href{https://github.com/tomsherborne/tram\_optimizer}{\texttt{github.com/tomsherborne/tram\_optimizer}}.}
    \item We highlight that TRAM empirically improves generalization for
    multiple out-of-distribution adaptation tasks across vision and natural
    language: cross-dataset adaptation for image classification, cross-domain
    language modeling and zero-shot cross-lingual transfer (\Cref{sec:results}).
    \item We analyze how TRAM limits catastrophic forgetting and optimizes
    flatter minima to improve fine-tuning. By characterizing major and minor
    distribution shifts, we identify how TRAM outperforms the trend in
    anticorrelated generalization scenarios. Our analysis verifies that TRAM
    optimizes a smoother loss surface for both in-domain and out-of-domain
    distributions. TRAM also improves representation similarity between seen and
    unseen distributions to improve cross-domain classification
    (\Cref{sec:results}).
\end{itemize}

\section{Background}\label{sec:background}

We describe SAM and trust region optimization, highlighting how these approaches
have similar goals. Our motivation for TRAM is the unifying features of each
approach outlined in \Cref{tab:sam_tr_comparison}.

\paragraphmini{Notation:}
We consider function $f:~X \rightarrow Y$ parameterized by weights $\theta$ and
evaluated by loss function $\ell:~Y\times Y\rightarrow\mathbb{R}_{+}$. The
expected loss on true distribution $\mathcal{D}$ is
$L_{\mathcal{D}}\left(\theta\right)=\mathbb{E}_{\left(x,y\right)\sim
\mathcal{D}}\left[\ell\left({y}, f\left(x; \theta\right)\right)\right]$ and the
empirical estimate is $L_{S}=\frac{1}{n}\sum_{S}\ell\left({y_i}, f\left(x_{i};
\theta\right)\right)$ sampling $n$ training samples, $S=\lbrace{({x}_{i},
{y}_{i})}\rbrace_{i=1}^{n}$, from $\mathcal{D}$. Functional distance on model
outputs is measured by the Kullback-Leibler divergence $\KL\left(p||q\right)$
between target $p$ and estimate $q$. We describe successful domain transfer to
distribution $\mathcal{D}'$ as a non-increasing loss for sample $S' \sim
\mathcal{D}'$.

\paragraphmini{Sharpness-Aware Minimization:}
\citet{foret2021sharpnessaware-SAM} define local sharpness as
$\max_{\lVert\epsilon\rVert_{2}\leq \rho}L_{S}\left(\theta + \epsilon\right) -
L_{S}\left(\theta\right)$. The SAM objective (\Cref{eq:sam_minmax}) regularizes
parameter magnitude to minimize this sharpness metric jointly with loss within
local parameter neighborhood $\rho$.

\noindent\begin{minipage}{.5\linewidth}
\begin{equation}
    L_{S}^{\rm SAM}=\min_{\theta}  \max_{\lVert\epsilon\rVert_2\leq \rho} L_{S}\left(\theta + \epsilon\right) + \frac{\lambda}{2}\lVert\theta\rVert_{2}^{2} \label{eq:sam_minmax}
\end{equation}
\end{minipage}%
\begin{minipage}{.5\linewidth}
\begin{equation}
  \epsilon^{\ast}_{\rm ASAM} = \rho
\frac{\theta^{2}\nabla L_{S}}{\lVert\theta \nabla L_{S}\rVert_{2}}  \label{eq:asam_rho}
\end{equation}
\end{minipage}

This min-max optimization problem is solved in alternating stages. Initial
ascent perturbs parameters $\theta$ to $\theta +\epsilon$, where $\epsilon$ is a
perturbation maximizing loss (to minimize local sharpness). The feasible region
for perturbation $\epsilon$ is a Euclidean spherical neighborhood with radius
$\rho>0$. Successive descent evaluates gradients at $\theta+\epsilon$ for
gradient descent at $\theta$ using the local worst-case loss. 

The optimal $\epsilon$, the perturbation for worst-case loss within the
$\rho$-ball, is the source of ongoing debate.
\citet{foret2021sharpnessaware-SAM} express a closed-form solution setting
$\epsilon$ as the radius $\rho$ scaled by the normalized gradient.
\citet{pmlr-v139-kwon21b-ASAM} propose Adaptive SAM (ASAM) to improve SAM with
invariance to the loss scaling. For ASAM, each parameter within $\theta$ is
perturbed by $\rho$ scaled by parameter gradient and the parameter norm
(\Cref{eq:asam_rho}). TRAM follows SAM in setting $\epsilon$ with scale
invariance and also augments $\epsilon$ such that the update in $\theta$
respects a maximum divergence in the function space.

\paragraphmini{Trust Region Regularization:} Trust region regularization
encourages low curvature during optimization by regularizing the function output
distribution with respect to a previous step's distribution. A fine-tuned model
with high curvature (i.e., distance) to pre-trained representations may struggle
to connect task-specific knowledge with novel domains. This approach proves
successful in penalizing large policy updates in reinforcement learning
\citep{pmlr-v37-schulman15-trust-region-policy}, encouraging local smoothness to
adversarial perturbation \citep{jiang-etal-2020-smart} and minimizing
catastrophic forgetting for domain transfer \citep{aghajanyan2021better}.

\Cref{eq:trust_region_1} defines the objective under Trust Region Policy
Optimization \citep[TRPO; ][]{pmlr-v37-schulman15-trust-region-policy}
constraining loss, $L_{S}$, with a regularization term $d_{\theta}$. TRPO
idealizes smoothness in $f\left(x\right)$ by regularizing local function
similarity to the previous iterate. The update at $t$ is constrained such that
changes in probability density, $p_{f}\left(\cdot|~x,\theta\right)$ are no
larger than some $\varepsilon$ measured by divergence
$d:~\mathcal{Y}\times\mathcal{Y}\rightarrow\mathbb{R}_{+}$. There are several
ways of defining $d$---we consider options in
\Crefrange{eq:trust_region_hist}{eq:trust_region_noise}.
\begin{equation}
        \min_{\theta} ~L_{S}\left(\theta \right)~ {\rm subject~to~} d_{\theta} \leq \varepsilon \label{eq:trust_region_1} 
\end{equation}
\Cref{eq:trust_region_hist} estimates the trust region as the KL divergence
between predictive distributions at the previous and current step. Intuitively,
penalizing divergence from prior steps encourages the function to stay ``close''
to the previous distribution i.e., within the trust region of equivalent output.
Across training, $d_{\theta}$ encourages small updates with low curvature
between fine-tuned and pre-trained models.
\begin{align}
    d_{\theta}\left(\theta_{t-1},\theta_{t}\right) &= \mathbb{E}_{x\sim D}\left[\KL\left(p_{f}\left(\cdot|~x,\theta_{t-1}\right)||p_{f}\left(\cdot|~x,\theta_{t}\right)\right)\right]
    \label{eq:trust_region_hist}
\end{align}
\Cref{eq:trust_region_noise} provides the penalty from R3F
\citep{aghajanyan2021better} where $d_{x}$ estimates the trust region by
sampling from inputs under parametric noise. This penalizes the divergence
between $p_{f}\left(\cdot|~x,\theta_{t}\right)$ and
$p_{f}\left(\cdot|~x+z,\theta_{t}\right)$ for some zero-mean noise
$z\sim\mathcal{N}\left(0,~\sigma^2\right)$. R3F proposes that sampling $z$
estimates the trust region by simulating a distribution shift in $p_{f}$
corresponding to perturbed $x+z$. This encourages similarity to a neighborhood
around $f\left(x,~\theta\right)$ with equivalent output. Either approach
estimates the permissible distance for an update in $\theta$ without increasing
local representation curvature. We focus on trust region methods to improve
generalization across distributions via improved leveraging of pre-trained
structure \interalia{jiang-etal-2020-smart}.
\begin{align}
    d_{x}\left(x+z, x\right)&=\mathbb{E}_{z\sim \mathcal{N}}\left[\KL\left(p_{f}\left(\cdot|~x+z,\theta\right)||p_{f}\left(\cdot|~x,\theta\right)\right)  \right]\label{eq:trust_region_noise}
\end{align}
\paragraphmini{Comparison:} SAM, TRPO, and R3F have similar goals in searching
for generalizable solutions while appearing superficially distinct. We compare
the broad motivations and qualities of methods in \Cref{tab:sam_tr_comparison},
highlighting both perspectives optimize for smoothness in different spaces.

SAM minimizes sharpness within a neighborhood in $\theta$ set by scalar
parameter $\rho$. Trust region regularization penalizes loss by scalar distance
$d_{\theta}~\text{or}~d_{x}$. We hypothesize that this regularization can inform
the size of the SAM neighborhood. Can we jointly minimize sharpness and penalize
high curvature in representations? Considering the sharpness objective
~$\left(\max_{\lVert\epsilon\rVert_{2}\leq \rho}L_{S}\left(\theta +
\epsilon\right) - L_{S}\left(\theta\right)\right)$~we consider if this
$\epsilon$ can also satisfy the \Cref{eq:trust_region_1} constraint of
$d_{\theta}~\text{or}~d_{x}<\varepsilon$. Our intuition here is to minimize
parameter sharpness (i.e., SAM) only within an update promoting low
representation curvature. Combining the features of these solutions could
improve generalization to unseen distributions during fine-tuning.

\begin{table}[t!]
\centering
\caption{Comparison between SAM-style, trust region and TRAM learning. SAM optimizes parameters for low sharpness, trust region methods optimize for low-curvature representations. TRAM combines these strategies to bound SAM-style learning within a trust region neighborhood.}
\label{tab:sam_tr_comparison}
\resizebox{\textwidth}{!}{%
\begin{tabular}{@{}ccccccc@{}}
\toprule
 &
  Goal &
  $\epsilon$ &
  Distance &
  Domain &
  Gradient &
  Forward/Backward \\ \midrule
SAM-style &
  Low-sharpness $\theta$ &
  \Cref{eq:asam_rho} &
  --- &
  $\rho$-ball &
  $\nabla L_{S}$ at $\theta+\epsilon$ &
  2 $\rightarrow$, 2 $\leftarrow$ \\
Trust region &
  Low-curvature $f\left(y|x, \theta\right)$ &
  --- &
  $d_{\theta}~\text{or}~d_{x}$ &
  $\KL$ over Distance &
  $\nabla L_{S} + d_{\theta}~\text{or}~d_{x}$ &
  2 $\rightarrow$, 1 $\leftarrow$ \\
TRAM &
  Both &
  \Cref{eq:tram_rho} &
  $d_{\theta}~\text{or}~d_{x}$ &
  $d_{\theta}$-$~\text{or}~d_{x}$-ball &
  $\nabla L_{S}$ at $\theta+\epsilon$ &
  3 $\rightarrow$, 2 $\leftarrow$ \\ \bottomrule
\end{tabular}%
}
\end{table}

\section{\textbf{TRAM}:~Trust Region Aware Minimization}\label{sec:tram}

We consider methods improving generalization by encouraging low-sharpness
parameters and task transfer by encouraging low curvature in representation
space. We introduce \textbf{TRAM}:~\textbf{T}rust \textbf{R}egion \textbf{A}ware
\textbf{M}inimization unifying sharpness-aware and trust region optimization.
\citet{pmlr-v162-kim22f-FisherSAM} raise that the $\rho$ hyperparameter defining
the ascent neighborhood in SAM is an ``ad hoc'' scaling with little relationship
to the loss landscape or parameter geometry. We propose to instead define the
ascent region by a trust region in representation space. 

TRAM substitutes $\rho$ in \Cref{eq:asam_rho} with the trust region metric,
$d:~\mathcal{Y}\times\mathcal{Y}\rightarrow\mathbb{R}^{\ast}_{+}$, as defined in
\Cref{sec:background}. We estimate trust regions using the divergence from a
prior model distribution ($d_{\theta}$,~\Cref{eq:trust_region_hist}) or
divergence from the current distribution under parametric noise
($d_{x}$,~\Cref{eq:trust_region_noise}). TRAM constrains the maximization domain
for ascent (i.e., $\theta\rightarrow\theta+\epsilon$) to the parameter corollary
for the trust region i.e., $\max_{\lVert\epsilon\rVert_2\leq d}$ substituted
within \Cref{eq:sam_minmax}. TRAM perturbs $\theta$ with a loss perturbation
only within the parameter neighborhood constrained for low representation
curvature. This introduces function curvature awareness within TRAM in addition
to the sharpness-awareness objective for flatter minima. In contrast, the
maximization region, $\rho$ in SAM/ASAM has no sensitivity to function
curvature. We build TRAM on ASAM, and not SAM, after observing strictly better
performance in our preliminary experiments.

\noindent\begin{minipage}{.5\linewidth}
\begin{equation}
    \nabla L_{\rm TRAM}\left(\theta\right) = \frac{\partial L_{S}}{\partial\theta}\biggr\rvert_{\theta=\theta+\epsilon^{\ast}_{\rm TRAM}} \label{eq:tram_final_grad}
\end{equation}
\end{minipage}
\begin{minipage}{.5\linewidth}
\begin{equation}
    \epsilon^{\ast}_{\rm TRAM} =
\frac{d~\theta^{2}\nabla L_{S}\left(\theta_{t}\right)}{\lVert\theta \nabla L_{S}\left(\theta_{t}\right)\rVert_{2}}  \label{eq:tram_rho}
\end{equation}
\end{minipage}%

The gradient descent update in TRAM is \Cref{eq:tram_final_grad}, where
$\epsilon^{\ast}_{\rm TRAM}$ is solved as \Cref{eq:tram_rho} by direct
substitution of $\rho$ in ASAM. \Cref{alg:tram} in \Cref{app:tram_fisher}
details the full training algorithm for TRAM based on the SAM-style min-max
optimization routine. TRAM does not require tuning a $\rho$ hyperparameter for
stable training. TRAM using $d_{\theta}$ introduces no new hyperparameters, and
using $d_{x}$ requires only tuning $\sigma$ for additive noise $z$. We
hypothesize that TRAM jointly minimizes parameter sharpness and representation
curvature to minimize catastrophic forgetting of pre-trained structure. Our
results in \Cref{sec:results} empirically validate this hypothesis.

\paragraphmini{Connection to ASAM:} 
The geometric interpretation of TRAM frames the maximization domain defined by
$d$ as a subdomain of the $\rho$-radius Euclidean ball defined in ASAM.\@
Whereas ASAM defines a fixed radius by $\rho$ at each step, TRAM instead uses
the nonzero $d$ radius constraining the maximization domain to additionally
satisfy the trust region constraint outlined in
\Crefrange{eq:trust_region_1}{eq:trust_region_noise}.~\citet[Theorem
2]{foret2021sharpnessaware-SAM} defines a PAC-Bayesian generalization bound for
SAM on $L_{\mathcal{D}}$ assuming $\rho>0$.~\citet[Theorem
3]{pmlr-v139-kwon21b-ASAM} identify a similarly valid bound when considering the
norm-adaptive scaling on $\epsilon^{\ast}_{\rm ASAM}$ as in \Cref{eq:asam_rho}.
We assume $d\leq\rho$ for similar asymptotic behavior for $\epsilon^{\ast}_{\rm
TRAM}$ to $\epsilon^{\ast}_{\rm ASAM}$. We infer that TRAM inherits the existing
generalization bound of ASAM for any $\rho>0$ directly substituted for $d$ i.e.,
TRAM is a subsolution of ASAM. We can constrain $d$ such that
$\max_{\theta_{\neg t}} d_{\theta}\left(\theta_{\neg
t},\theta_{t}\right)\leq\rho$ or $\max_{z}d_{x}\left(x+z, x\right)\leq\rho$ to
enforce this bound $d\in\left.\left(0,\rho\right.\right]$. However, we
empirically observe this constraint is satisfied for the optimal setting of
$\rho$ in ASAM.\@

\paragraphmini{Improving Efficiency with TRAM-Fisher:}
\citet{pmlr-v162-kim22f-FisherSAM} propose an alternative to SAM removing the
Euclidean assumption for parameter geometry. Fisher SAM (FSAM) instead exploits
the \emph{statistical manifold} induced by the Fisher Information metric of
predictive distribution of the function, $p_{f}\left(y|~x,\theta\right)$
\citep{amari-98-natural-gradient} to set $\epsilon$. This measures statistical
divergence between $\theta$ and $\theta+\epsilon$ resulting in $\epsilon_{\rm
FSAM}^{\ast}$ in \Cref{eq:fsam_rho} defining an ellipsoid around $\theta$ scaled
by the Fisher Information matrix, $F\left(\theta\right)$. $F\left(\theta\right)$
is prohibitively expensive at scale and is approximated with
\Cref{eq:samfisher_fhat}, the diagonal of the squared gradient sum for each
batch $B$.

\noindent\begin{minipage}{.43\linewidth}
\begin{equation}
\epsilon_{\rm FSAM}^{\ast}= \frac{F{\left(\theta \right)}^{-1}\nabla L_{S}}{\sqrt{\nabla L_{S}F{\left(\theta \right)}^{-1}\nabla L_{S}}} \label{eq:fsam_rho}
\end{equation}
\end{minipage}%
\begin{minipage}{.57\linewidth}
\begin{equation}
\hat{F}\left(\theta\right) = {\rm Diag}{\left(\frac{1}{|B|} \sum_{i\in B} \left(\log p_{f}{\left(y_{i}|x_{i}, \theta\right)}\right)\right)}^{2} \label{eq:samfisher_fhat}
\end{equation}
\end{minipage}

We propose TRAM-Fisher as an efficient variant of TRAM inspired by Fisher SAM.\@
Where FSAM measures the Fisher Information geometry of $\theta$ under input $x$,
we instead sample the geometry of $\theta$ under the trust region estimation
from $x+z$. Our proposal is minimal: replace $p\left(y_{i}|x_{i}, \theta\right)$
with $p\left(y_{i}|x_{i}+z_{i}, \theta\right)$ to estimate the Fisher
Information Matrix of the \emph{trust region} neighborhood as
$\mathbb{E}_{z\sim\mathcal{N}}\left[\hat{F}\left(x+z; \theta\right)\right]$. We
sample parametric noise ${\{z_{i}\}}_{i=0}^{|B|}$ identically to TRAM and now
scale learning with the information geometry of the low curvature neighborhood,
$f\left(x+z\right)$. TRAM-Fisher uses the same number of forward/backward passes
as FSAM and only requires additional processing to sample $z$ and compute $x+z$.
TRAM-Fisher matches FSAM in runtime efficiency (with marginal additional
operations) and performs competitively across our experiments. The full
TRAM-Fisher algorithm is shown in \Cref{app:tram_fisher}.

\paragraphmini{Summary:} We propose three variants of TRAM, and TRAM-Fisher,
summarized in \Cref{tab:tram_types}. TRAM-${\theta_{t-1}}$ follows TRPO
\citep{pmlr-v37-schulman15-trust-region-policy} in using previous step
parameters, $\theta_{t-1}$, to measure the trust region. We also propose a
simplification of TRAM-${\theta_{t-1}}$ estimating the trust region using
$d_{\theta}$ between current $\theta_{t}$ and pre-trained model $\theta_{0}$.
TRAM-${\theta_{0}}$ improves training efficiency by removing an updating
$\theta_{t-1}$ state. TRAM-${x}$ follows R3F \citep{aghajanyan2021better} using
noise-based trust region measurement with additional hyperparameter $z$ for
sampling parametric noise. Practically, TRAM requires one additional forward
pass adding marginal overhead to the extant complexity of SAM-style training.
Despite this additional cost, \Cref{sec:results} identifies empirical benefits
to TRAM and targeted improvement to out-of-domain loss surface sharpness and
cross-domain representation similarity.

\begin{table}[t]
\centering
\caption{We propose four variants of TRAM based on different trust region
estimations. TRAM-$\theta_{t-1}$ uses divergence against the previous step;
TRAM-$\theta_{0}$ is a simplifying heuristic of this divergence against the
pre-trained model only; and TRAM-$x$ uses noised input divergence, $d_{x}$.
TRAM-Fisher extends FSAM by measuring the Fisher Information metric around the
trust region.} \resizebox{0.9\textwidth}{!}{%
\begin{tabular}{@{}lcccc@{}}
\toprule
Variant             & Trust region measurement                          & $\epsilon$         & Domain            & Forward/Backward                \\ \midrule
TRAM-${\theta_{t-1}}$ & $d_{\theta}\left(\theta_{t-1}, \theta_{t}\right)$ & \Cref{eq:tram_rho} & $d_{\theta}$-ball & 3 $\rightarrow$, 2 $\leftarrow$ \\
TRAM-${\theta_{0}}$   & $d_{\theta}\left(\theta_{0}, \theta_{t}\right)$   & \Cref{eq:tram_rho} & $d_{\theta}$-ball & 3 $\rightarrow$, 2 $\leftarrow$ \\
TRAM-${x}$ & $d_{x}\left(x+z, x\right), z\sim \mathcal{N}\left(0,
  \sigma^{2}\right)$ & \Cref{eq:tram_rho} & $d_{x}$-ball & 3 $\rightarrow$, 2
  $\leftarrow$ \\
TRAM-Fisher &
  $\hat{F}\left(x+z; \theta\right), z\sim \mathcal{N}\left(0, \sigma^{2}\right)$ &
  \Cref{eq:fsam_rho} &
  $\hat{F}$-ellipse &
  2 $\rightarrow$, 2 $\leftarrow$ \\ \bottomrule
\end{tabular}%
}
\label{tab:tram_types}
\end{table}

We outline our datasets in \Cref{app:data}, and experiment design in
\Cref{app:experiments} for both vision and language modalities. We compare to
gradient descent methods (SGD, Adam), sharpness aware methods (SAM, ASAM, FSAM),
and trust region methods (TRPO, R3F, MESA) further detailed in
\Cref{app:experiments:design:baselines}. Broadly, we investigate the hypothesis
that \textit{out-of-distribution generalization improves by jointly minimizing
parameter sharpness and representation curvature in the function.}

\section{Results}
\label{sec:results}

\subsection{Cross-dataset image classification}\label{results:image}

\begin{table}[th]
    \centering
    \caption{Cross-dataset adaptation from ImageNet to CIFAR-100, Stanford Cars
        and Oxford Flowers. We report Top-1 classification accuracy averaged
        over five runs, $\pm$ the 95\% confidence interval, for direct
        comparison to \citet{pmlr-v162-kim22f-FisherSAM}.
        }
    \label{tab:viz}
    \small
    \begin{tabular}{@{}lccc@{}}
    \toprule
    & CIFAR-100 $\left(\uparrow\right)$      & Cars $\left(\uparrow\right)$          & Flowers $\left(\uparrow\right)$       \\ \midrule
    SGD                   & \colmin{87.97}$\pm$0.12 & \colmin{92.85}$\pm$0.31 & \colmin{94.53}$\pm$0.20 \\
    SAM                   & 87.99$\pm$0.09 & 93.29$\pm$0.01 & 95.05$\pm$0.06 \\
    ASAM                  & \colmin{87.97}$\pm$0.08 & 93.28$\pm$0.02 & 95.08$\pm$0.10 \\
    FSAM                  & 88.39$\pm$0.13 & 93.42$\pm$0.01 & 95.26$\pm$0.03 \\ \midrule
    TRAM-${\theta_{t-1}}$ & 88.47$\pm$0.16 & \colmax{93.49}$\pm$0.04 & \colmax{97.07}$\pm$0.10 \\
    TRAM-${\theta_0}$     & 88.31$\pm$0.09 & 93.16$\pm$0.07 & 95.53$\pm$0.10 \\
    TRAM-${x}$            & \colmax{88.78}$\pm$0.01 & 93.32$\pm$0.11 & 96.34$\pm$0.03 \\
    TRAM-Fisher           & 88.02$\pm$0.18 & 93.12$\pm$0.13 & 94.90$\pm$0.11  \\ \bottomrule
    \end{tabular}%
\end{table}

First, we validate the performance of TRAM in a standardized setting for
comparison to other SAM-style optimizers. We evaluate adapting ViT-base
\citep{dosovitskiy2021an-vit} from ImageNet pre-training to image classification
fine-tuning. We follow the setup of \citet[Section
5.1]{pmlr-v162-kim22f-FisherSAM} evaluating adaptation to CIFAR-100
\citep{krizhevsky2009learning-cifar}, Stanford Cars
\citep{krause-2013-stanford-cars}, and Oxford Flowers
\citep{nilsback-2008-flowers}. \Cref{app:experiments:design:viz} details our
experiment design.

\Cref{tab:viz} details the Top-1 accuracy results for this experiment with
direct comparison to \citet[Table 3]{pmlr-v162-kim22f-FisherSAM}. The
best-performing variant of TRAM (TRAM-$\theta_{t-1}$ or TRAM-$x$) is
significantly superior to the closest FSAM competitor ($p<0.01$). Other variants
of TRAM, TRAM-$\theta_{0}$ or TRAM-Fisher, are largely competitive with prior
methods. Our observations validate the hypothesis that TRAM improves adaptation
across datasets during fine-tuning for image classification. This comparison
acts as a sanity check and demonstrates the utility of our method compared to
other SAM-style optimizers. TRAM yields improved fine-tuned image classification
models by encouraging smoothness in parameter and function space.

\subsection{Cross-domain language modeling}\label{results:lm}

\begin{table}[t]
  \caption{M2D2 perplexity (lower is better) on Wikipedia (upper) \& S2ORC
  (lower) splits. TRAM-${\theta_{t-1}}$ significantly improves over prior work
  ($p<0.01$ Kolmogorov-Smirnov test). Results are grouped as: (i) optimizers;
  (ii) trust region methods; and (iii) TRAM variants. The leftmost column is the
  training domain and we evaluate zero-shot perplexity on ten domains unseen
  during fine-tuning (full details in \Cref{app:data}). {\sc ZS Avg.} is the
  macro-average of all zero-shot domains.}
  \small
  \begin{subtable}[v]{\textwidth}
  \resizebox{\textwidth}{!}{%
    \begin{tabular}{@{}lc|cccccccccc|c@{}} 
      \toprule
    \textbf{Wiki} & {\sc Soc.} & {\sc Cult.} & {\sc Gen.} & {\sc Health.} & {\sc Hist.} & {\sc Human.} & {\sc Math.} & {\sc Nat.} & {\sc Phil.} & {\sc Rel} & {\sc Tech.} & {\sc ZS Avg.} $\downarrow$ \\ \midrule
    GPT-2 & 27.2 & 27.7 & 27.8 & 24.5 & 29.2 & 28.8 & 28.6 & 29.4 & 27.8 & 27.7 & 28.7 & 28.0 \\  \midrule
    Adam & \colmin{24.8} & \colmin{26.3} & \colmin{26.4} & \colmin{23.6} & \colmin{27.2} & \colmin{27.0} & \colmin{27.4} & \colmin{27.6} & \colmin{26.3} & \colmin{25.8} & {27.4} & \colmin{26.5} \\
    SAM & 24.5 & 25.9 & 26.0 & 23.1 & 26.9 & 26.6 & 26.6 & 27.2 & 25.8 & 25.5 & 27.0 & 26.1 \\
    ASAM & \colmin{24.8} & 25.4 & 25.6 & 22.5 & 27.1 & 26.4 & 26.3 & 26.7 & 25.5 & 25.5 & \colmin{28.1} & 25.9 \\
    FSAM & 21.7 & 23.0 & 23.3 & 20.6 & 23.9 & 23.7 & 23.8 & 24.0 & 23.1 & 22.8 & 24.0 & 23.2 \\  \midrule
    TRPO & 21.8 & 23.0 & 23.3 & 20.7 & 24.0 & 23.7 & 23.8 & 24.0 & 23.1 & 22.8 & 24.1 & 23.3 \\
    R3F & 21.8 & 23.0 & 23.3 & 20.7 & 24.0 & 23.7 & 23.8 & 24.0 & 23.1 & 22.8 & 24.1 & 23.3 \\
    MESA & 23.1 & 24.0 & 24.3 & 21.5 & 25.4 & 24.9 & 24.8 & 25.2 & 24.1 & 24.0 & 25.1 & 24.3 \\  \midrule
    TRAM-${x}$ & 21.9 & 23.1 & 23.4 & 20.7 & 24.0 & 23.3 & 23.9 & 23.9 & 23.2 & 22.7 & 23.9 & 23.2 \\
    TRAM-${\theta_{t-1}}$ & \colmax{20.9} & \colmax{22.4} & \colmax{22.7} & \colmax{20.1} & \colmax{23.1} & \colmax{22.9} & \colmax{23.2} & \colmax{23.3} & \colmax{22.4} & \colmax{22.0} & \colmax{23.4} & \colmax{22.5} \\
    TRAM-${\theta_{0}}$ & 21.9 & 23.1 & 23.4 & 20.7 & 23.9 & 23.3 & 23.9 & 23.8 & 23.1 & 22.7 & 23.9 & 23.2 \\
    TRAM-Fisher & 22.5 & 23.7 & 24.0 & 21.3 & 24.6 & 24.0 & 24.7 & 24.6 & 23.8 & 23.3 & 24.6 & 23.9 \\ \bottomrule
    \end{tabular}
    } 
    \label{tab:m2d2:wiki}
    \vspace{0.5em}
  \end{subtable}
  
  \begin{subtable}[v]{\textwidth}
  \centering
    \resizebox{\textwidth}{!}{%
    \begin{tabular}{@{}lc|cccccccccc|c@{}}
    \toprule
  \textbf{S2ORC} &
    {\sc Math} &
    {\sc Art} &
    {\sc Astro} &
    {\sc CondM.} &
    {\sc CS} &
    {\sc Econ.} &
    {\sc NLin.} &
    {\sc Phil.} &
    {\sc Phys.} &
    {\sc QBio} &
    {\sc Stat} &
    {\sc ZS Avg.} $\downarrow$ \\ \midrule
  GPT-2   & 27.6 & 35.8    & 32.4    & 30.9 & 27.9 & 29.5 & 27.6 & 33.7    & 33.5 & 30.9 & 23.4 & 30.6 \\ \midrule
  Adam    & 11.4 & 44.2    & 33.9    & 20.1 & 21.2 & 21.0 & 14.7 & 41.9    & 29.5 & 30.8 & 16.9 & 27.4 \\
  SAM     & 10.5 & 45.3    & 33.2    & 18.7 & 20.3 & 20.0 & 13.7 & 42.4    & 28.3 & 30.2 & 16.1 & 26.8 \\
  ASAM    & 10.3 & 45.6    & 33.2    & 18.5 & 20.1 & 19.8 & 13.5 & 42.6    & 28.2 & 30.2 & 15.9 & 26.8 \\
  FSAM    & 10.4 & 45.6    & 33.3    & 18.5 & 20.2 & 19.9 & 13.5 & 42.7    & 28.3 & 30.2 & 15.9 & 26.8 \\ \midrule
  TRPO    & 10.4 & 46.0    & 33.4    & 18.6 & 20.3 & 20.0 & 13.6 & 42.9    & 28.4 & 30.4 & 16.0 & 26.9 \\
  R3F     & 10.4 & 46.0    & 33.4    & 18.6 & 20.2 & 20.0 & 13.6 & 42.9    & 28.4 & 30.4 & 16.0 & 26.9 \\
  MESA    & \colmin{11.9} & \colmax{44.1} & \colmin{34.1} & \colmin{20.8} & \colmin{21.7} & \colmin{21.6} & \colmin{15.3} & \colmax{41.7} & \colmin{30.0} & \colmin{31.0} & \colmin{17.4} & \colmin{27.8} \\ \midrule
  TRAM-${x}$    & 10.4 & 44.9    & 33.0    & 18.6 & 20.1 & 19.9 & 13.6 & 42.0    & 28.1 & 30.0 & 15.9 & 26.6 \\
  TRAM-${\theta_{t-1}}$ & \colmax{9.6} & \colmin{46.8} & 32.5 & \colmax{17.2} & \colmax{19.2} & \colmax{18.9} & \colmax{12.6} & \colmin{43.3} & \colmax{27.0} & \colmax{29.6} & \colmax{15.0} & \colmax{26.2} \\
  TRAM-${\theta_{0}}$ & 10.4 & 44.8    & 33.0    & 18.6 & 20.1 & 19.9 & 13.6 & 42.0    & 28.2 & 30.0 & 15.9 & 26.6 \\
  TRAM-Fisher  & 10.5 & {46.1}    & \colmax{32.4} & 18.7 & 20.3 & 20.0 & 13.6 & 43.0    & 28.2 & 30.3 & 16.0 & 26.9 \\ \bottomrule
  \end{tabular}
    } 
    \label{tab:m2d2:s2orc}
   \end{subtable}
   \label{tab:m2d2}
   \vspace{-2em}
\end{table}

We now consider zero-shot cross-domain language modeling using the M2D2 Corpus
\citep{reid-etal-2022-m2d2} outlined in \Cref{app:data}. We hypothesize that
TRAM can improve domain transfer in language modeling by retaining
domain-agnostic information from pre-training when fine-tuning to a specific
domain. We train a GPT-2 Base model \citep{radford2019language-gpt2} on the
largest domain in each split of M2D2 ({\sc Soc.} domain 379M tokens for
Wikipedia and {\sc Math} 1.4B tokens for S2ORC) and evaluate perplexity across
ten domains unseen during fine-tuning. \Cref{app:experiments:design:lm} details
our complete experiment design. 

Our results in \Cref{tab:m2d2} validate our hypothesis for TRAM in the
cross-domain setting to improve out-of-domain language modeling fine-tuning on a
single domain. All TRAM variants (excluding TRAM-Fisher) perform comparably or
above competitors in zero-shot transfer across both splits of M2D2. TRAM
improves domain transfer in fine-tuned models by better leveraging pre-trained
information from unseen domains within the smoother minima idealized by
SAM-style training. Generally, the naive Adam baseline or the MESA trust region
comparison perform poorest at cross-domain language modeling for Wikipedia or
S2ORC splits respectively. As with image classification, FSAM is the strongest
competitor to TRAM. The best variant in both splits is TRAM-${\theta_{t-1}}$
improving in-domain and average zero-shot perplexity. TRAM-${\theta_{t-1}}$ uses
the TRPO method of estimating the trust region using the parameters of the
previous step. This variant always yields the lowest perplexity in the training
domain and the majority of similar and distant zero-shot domains. We
additionally verify that TRAM performs competitively at a larger model scale
using GPT2-XL (1.5B parameters) in \Cref{tab:m2d2:s2orc:xl} in \Cref{app:lm-xl}.
We also compare against a naive combination of methods (e.g., ASAM+TRPO) in
\Cref{app:combo}.

TRAM improves perplexity for all domains in the Wikipedia split, where all
zero-shot domains are positively correlated with the training domain perplexity.
However, we observe that perplexity \emph{degrades} for domains distant from the
fine-tuning domain in S2ORC ({\sc Math}) which benefit less from shared
features. Given that neither SAM-style nor trust region methods inverted this
anticorrelation trend, it is unsurprising that TRAM follows suit. This
confounder results in the overall best model, TRAM-$\theta_{t-1}$, reporting the
\textit{worst} performance for the distant domains where the overall poorest
model, MESA, reports the \textit{best} performance. We suggest that optimization
alone may be insufficient to improve zero-shot domain adaptation for larger
distribution shifts. We discuss further the correlation between domain-specific
perplexity in \Cref{app:domain_correlation}.

\subsubsection{Easy and hard
generalization}\label{results:analysis:easy_and_hard}

When evaluating performance variation between different distributional
shifts---we find that TRAM improves on all prior work for minor shifts (e.g.,
{\sc Math} to Physics/{\sc Phys.}) and generally matches or improves on a
negative trend for major shifts (e.g., {\sc Math} to {\sc Art}). Discussion of
out-of-domain generalization often overlooks differences between major and minor
shifts. In practice, in-domain performance has a very different relationship to
performance when generalizing to a major domain shift rather than a minor shift.
Considering minor distribution shifts, accuracy is strongly correlated on
in-domain and out-of-domain datasets \citep{miller2021accuracy}. However, major
distribution shifts may lead to scenarios where performance is instead
\textit{anticorrelated} with in-domain accuracy \citep{Teney2022IDAO}.
Considering these scenarios in the S2ORC task, we observe that models trained
using TRAM often perform better on new domains than their in-domain performance
would predict. Furthermore, TRAM improves perplexity across both minor and major
distribution shifts.

\begin{figure}
    \centering
    {\phantomsubcaption\label{fig:near-domains}
    \phantomsubcaption\label{fig:art} \phantomsubcaption\label{fig:philosophy}}
    \caption{Perplexity on S2ORC training domain ({\sc Math}) and zero-shot
    domains. We report perplexity across: (\subref{fig:near-domains}) domains
    correlated with {\sc Math} as \textsc{Stem} domains (see
    \Cref{app:domain_correlation}), (\subref{fig:art}) {\sc Art} domain, and
    (\subref{fig:philosophy}) the Philosophy ({\sc Phil.}) domain. Each figure
    includes linear regression trends: the \textcolor{blue}{blue} dotted trend
    is for prior work and \textcolor{green}{green} dashed line includes all TRAM
    variants. Positive slope ($\rho>0$) represents correlated domains, negative
    slope ($\rho<0$) represents anticorrelated domains. We report Pearson $\rho$
    correlation for the \textcolor{blue}{blue} trend noting $p<0.01$
    significance.}
    \includegraphics[width=\textwidth]{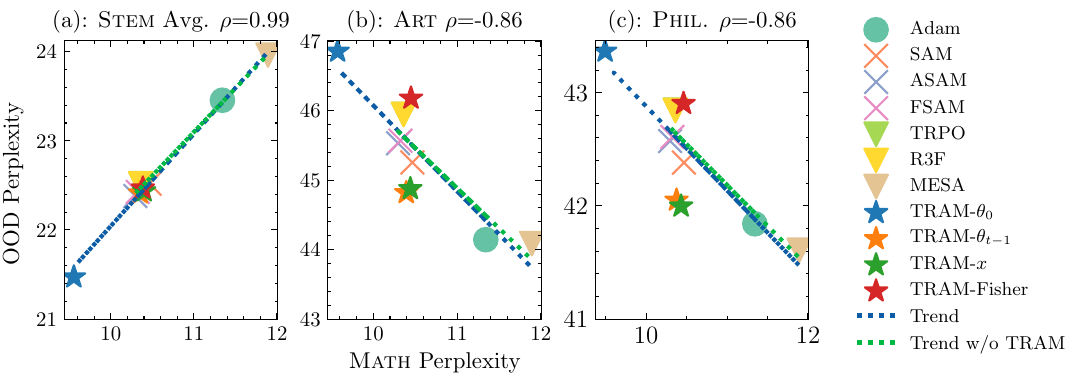}
    \label{fig:distant-domains}
    \vspace{-3em}
\end{figure}

\Cref{fig:near-domains} shows the close positive correlation between performance
on the training domain ({\sc Math}) and the average across all other STEM
disciplines, considering all optimization approaches. As detailed in
\Cref{app:domain_correlation}, performance correlates with $\rho > 0.8$ between
{\sc Math} and each individual STEM category. Considering the
\textcolor{blue}{blue} dotted trend for previous optimization methods (excluding
TRAM), we see that all TRAM optimizers fall on or marginally below the line.
This result suggests that TRAM not only supports in-domain performance but
specifically improves generalization to similar domains.

By contrast, we find there is generally a trade-off between performance on {\sc
Math} and the hardest anticorrelated domains: {\sc Art} (\Cref{fig:art}) and
Philosophy ({\sc Phil}, \Cref{fig:philosophy}). Both TRAM-${x}$ and
TRAM-${\theta_{0}}$ fall far below the trend for previous algorithms where
in-domain improvement worsens out-of-domain perplexity. TRAM not only matches or
outperforms existing methods on easier generalization cases, but exhibits a
lesser trade-off between easy and hard generalization compared to all previous
approaches.

\subsection{Zero-shot cross-lingual transfer}\label{results:xlt}

Finally, we now consider if TRAM improves cross-lingual adaptation during
monolingual fine-tuning. We adapt a multilingual pre-trained model to an English
entailment classification task (NLI) and then evaluate the zero-shot
cross-lingual capability for the model to classify entailment from inputs in 14
unseen languages. We hypothesize that TRAM benefits cross-lingual transfer via
improved application of multilingual pre-trained information to a task with only
English training data. In general, languages closer to English (e.g., French,
German) are ``easier'' for transfer than distant or low-resource languages
(e.g., Urdu, Swahili) \citep{ahmad-etal-2019-difficulties}. An ideal system will
produce equivalent cross-lingual transfer for all zero-shot languages. Our
complete experiment design is outlined in \Cref{app:experiments:design:xlt}. We
train an XLM-Roberta-based model \citep{conneau-etal-2020-unsupervised} on
English MultiNLI \citep{williams-etal-2018-broad} and report accuracy results
for the XNLI cross-lingual entailment benchmark (discussed in \Cref{app:data}).

\Cref{tab:xnli} highlights that TRAM improves over all competing methods for the
cross-lingual transfer objective, similar to our findings for cross-dataset
image classification and cross-domain language modeling. Similar to the above
tasks, TRAM-${x}$ and TRAM-${\theta_{t-1}}$ are the best-performing algorithms
reporting both the strongest in-domain and average out-of-domain accuracy.
Either TRAM variant is the best method across all individual languages. We
identify that all methods worsen for languages distant from English in a similar
trend to language modeling for anticorrelated domains. However, here TRAM is
strictly superior to any other method for both near and distant languages to
English. Notably, TRAM-Fisher significantly improves upon FSAM ($p<0.01$)
despite the close similarity in methods. Given the additional forward pass
required for TRAM-${x}$, TRAM-Fisher represents a better performance-complexity
trade-off which is competitive in some tasks. We analyze the loss surface and
representation transfer in \Cref{tab:analysis_total} to verify that TRAM extends
a low-curvature loss surface and representation smoothness to all zero-shot
languages. In \Cref{app:xnli_distance_metric}, we train a model using TRAM with
alternative distances for trust region measurement to analyze the criticality of
using KL divergence. We observe that TRAM is robust to multiple distances with
marginal degradation. These results empirically verify our hypothesis that
training with complementary SAM-style and trust region methods improves the
language transferability of a fine-tuned model. 

\begin{table}[t]
\centering
\caption{XNLI accuracy (higher is better) for training language ({\sc En}) and
14 zero-shot target languages summarised by {\sc ZS Avg.} (key in
\Cref{app:data}). All TRAM variants significantly outperform other methods
($p<0.01$ Wilcoxon test). Results are grouped as: (i) optimizers; (ii) trust
region methods; and (ii) TRAM variants. We report the mean across 20 seeds with
standard deviation in \Cref{tab:xnli_stddev}.} 

\resizebox{\textwidth}{!}{%
\begin{tabular}{@{}lc|cccccccccccccc|c@{}}
\toprule
 &
  {\sc En} &
  {\sc Ar} &
  {\sc Bg} &
  {\sc De} &
  {\sc El} &
  {\sc Es} &
  {\sc Fr} &
  {\sc Hi} &
  {\sc Ru} &
  {\sc Sw} &
  {\sc Th} &
  {\sc Tr} &
  {\sc Ur} &
  {\sc Vi} &
  {\sc Zh} &
  {\sc ZS Avg.} $\uparrow$ \\ \midrule
Adam &
  \colmin{83.9} &
  \colmin{71.2} &
  \colmin{77.1} &
  \colmin{75.7} &
  \colmin{75.2} &
  \colmin{78.3} &
  77.6 &
  69.6 &
  \colmin{74.9} &
  64.6 &
  \colmin{71.2} &
  \colmin{72.2} &
  65.8 &
  74.1 &
  \colmin{73.1} &
  \colmin{72.9} \\
SAM                   & 84.8 & 72.1 & 78.1 & 76.7 & 75.7 & 79.0 & 77.9       & 69.8       & 75.7 & 65.2       & 71.8 & 73.1 & 66.8       & 75.1       & 74.2 & 73.7 \\
ASAM                  & 85.0 & 72.0 & 78.4 & 76.9 & 76.1 & 79.5 & 78.5       & 70.4       & 76.1 & 65.2       & 72.5 & 73.4 & 66.9       & 75.5       & 74.2 & 74.0 \\
FSAM                  & 84.7 & 72.2 & 78.1 & 76.9 & 76.0 & 79.3 & 78.4       & 70.0       & 76.1 & 65.1       & 72.2 & 73.0 & 66.8       & 75.3       & 74.2 & 73.8 \\ \midrule
TRPO                  & 84.9 & 71.3 & 77.7 & 76.2 & 75.3 & 78.6 & \colmin{77.3} & \colmin{69.2} & 75.2 & 64.4       & 71.6 & 72.4 & \colmin{65.3} & \colmin{73.8} & 73.3 & 73.0 \\
R3F                   & 85.5 & 72.7 & 78.9 & 77.5 & 76.8 & 79.9 & 79.2       & 70.7       & 76.8 & 66.2       & 72.9 & 73.9 & 66.6       & 75.8       & 74.6 & 74.5 \\
MESA                  & 84.9 & 71.9 & 77.9 & 76.7 & 75.7 & 78.8 & 77.8       & 69.6       & 75.8 & \colmin{64.1} & 72.1 & 72.4 & 65.7       & 74.4       & 73.9 & 73.3 \\ \midrule
TRAM-${x}$ &
  \colmax{86.2} &
  \colmax{73.5} &
  \colmax{79.8} &
  \colmax{78.3} &
  \colmax{77.5} &
  \colmax{80.9} &
  79.6 &
  71.4 &
  \colmax{77.5} &
  66.0 &
  \colmax{73.8} &
  \colmax{74.3} &
  \colmax{67.6} &
  \colmax{76.7} &
  \colmax{75.9} &
  \colmax{75.2} \\
TRAM-${\theta_{t-1}}$ &
  \colmax{86.2} &
  73.1 &
  79.5 &
  78.2 &
  77.0 &
  80.2 &
  \colmax{79.7} &
  \colmax{71.5} &
  \colmax{77.5} &
  \colmax{66.4} &
  73.3 &
  74.2 &
  67.5 &
  \colmax{76.7} &
  75.8 &
  75.0 \\
TRAM-${\theta_{0}}$ & 85.6 & 72.9 & 79.3 & 77.8 & 77.4 & 80.2 & 79.6       & 71.2       & 77.1 & 65.9       & 73.3 & 74.2 & 67.5       & \colmax{76.7} & 75.8 & 74.9 \\
TRAM-Fisher           & 84.3 & 73.1 & 78.7 & 77.1 & 76.2 & 79.5 & 78.4       & 71.4       & 76.6 & 65.7       & 73.2 & 73.6 & 67.5       & 75.5       & 75.5 & 74.4 \\ \bottomrule
\end{tabular}%
}
\label{tab:xnli}
\vspace{-1em}
\end{table}

\paragraphmini{Loss surface dynamics:}
Investigating the loss surface, we test the hypothesis that TRAM leads to
flatter minima on both \emph{in-domain} and \emph{out-of-domain} data. We
evaluate validation set $\epsilon$-sharpness \citep{keskar2017on}, defined in
\Cref{app:sharpness_def}, across 20 trained models. We report in-domain (for
English) and out-of-domain (zero-shot languages) $\epsilon$-sharpness in
\Cref{tab:analysis:sharpness} across TRAM and baselines (omitting models which
under-performed). Most methods unsurprisingly demonstrate a lower in-domain
sharpness but poorer out-of-domain sharpness. TRAM yields a smoother solution
for both the in-domain and out-of-domain regions of the loss surface.  We also
observe an improved average Pearson correlation (and lower variance) between
in-distribution and out-of-distribution sharpness using TRAM. This infers that
the relationship between loss surfaces of different distributions is more
desirably predictable with TRAM. Notably, other SAM-style methods are
\emph{worse than Adam} for out-of-domain sharpness---suggesting that current SAM
algorithms (excluding TRAM) are possibly ``sharpness-aware'' only within the
training distribution.

\begin{table}[t]
    \centering
    \caption{Analysis of (a) $\epsilon$-sharpness and (b) CKA representation
            similarity for TRAM. We measure each metric using the XNLI
            validation set and report for the training language ({\sc En}) and
            the zero-shot languages ({\sc ZS}). We report mean of 20 runs $\pm$
            standard deviation across languages and the Pearson correlation
            between {\sc En} and {\sc ZS Avg.} $\epsilon$-sharpness across runs.
            }
    \vspace{-0.5em}
    \footnotesize
    \begin{subtable}[h]{0.5\textwidth}
        \centering
         \begin{tabular}{@{}lccc@{}} \toprule
            (a)~$\epsilon$-sharpness $\downarrow$ & {\sc En} & {\sc ZS Avg.} & Pearson $\rho$ \\ \midrule
            Adam & 2.16 & 1.98$\pm$~0.79 & 0.29$\pm$0.20 \\
            SAM & 1.43 & 3.32$\pm$~0.96 & 0.26$\pm$0.34 \\
            ASAM & 2.57 & 2.22$\pm$~0.79 & 0.38$\pm$0.12 \\
            FSAM & 2.34 & 2.62$\pm$~0.29 & 0.27$\pm$0.71 \\
            TRPO & 6.17 & 2.36$\pm$~1.02 & 0.52$\pm$0.25 \\
            R3F & 6.22 & 2.56$\pm$~1.21 & 0.50$\pm$0.12 \\
            MESA & 2.76 & 5.48$\pm$~0.75 & 0.21$\pm$0.25 \\ \midrule
            TRAM-${\theta_{t-1}}$ & \colmax{0.50} & \colmax{1.19}$\pm$0.38 & 0.60$\pm$0.15 \\
            TRAM-${\theta_{0}}$ &  0.75 & 1.92$\pm$0.24 & {{0.58}$\pm$0.27} \\
            TRAM-${x}$ & 0.61 & {1.49}$\pm$~0.49 & \colmax{0.75}$\pm$0.18 \\  
            TRAM-Fisher & 1.67 & 2.02$\pm$~0.40 & {{0.42}$\pm$0.37}  \\ \bottomrule
            \end{tabular}
        \label{tab:analysis:sharpness}
    \end{subtable}
    \hspace{1em}
    \begin{subtable}[h]{0.3\textwidth}
    \centering
        \begin{tabular}{@{}lcc@{}} \toprule
(b)~CKA $\uparrow$                   & {\sc En}      & {\sc ZS Avg.}               \\ \midrule
Adam                    & 0.69          & 0.44$\pm$~0.10          \\
SAM                     & 0.69          & 0.42$\pm$~0.10          \\
ASAM                    & 0.69          & 0.42$\pm$~0.10          \\
FSAM                    & 0.73          & 0.48$\pm$~0.10          \\
TRPO                    & 0.70           & 0.45$\pm$~0.10          \\
R3F                     & 0.66          & 0.40$\pm$~0.10           \\
MESA                    & 0.67          & 0.42$\pm$~0.10          \\ \midrule
TRAM-${\theta_{t-1}}$ & \colmax{0.77} & \colmax{0.57}$\pm$~0.10 \\
TRAM-${\theta_{0}}$ &  0.69 & 0.45$\pm$~0.11 \\
TRAM-${x}$            & 0.75          & 0.54$\pm$~0.11         \\
TRAM-Fisher &  0.72 & 0.49$\pm$~0.10 \\
\bottomrule
\end{tabular}
        \label{tab:analysis:representation}
    \end{subtable}
    \label{tab:analysis_total}
\end{table}
\paragraphmini{Representation transfer:}
We analyze the similarity of pre-trained and fine-tuned representations for the
same setup of XNLI. We hypothesize that if TRAM optimizes within the trust
region, pre- and post-fine-tuned representations will be more similar to allow
better usage of pre-trained structure. We measure this relationship using CKA
similarity \citep{pmlr-v97-kornblith19a} defined in \Cref{app:cka_def}. Similar
to the previous analysis, we observe that TRAM produces representations that are
more similar to pre-trained XLM-Roberta representations than any competitor.
This applies to both the {\sc En} case and the {\sc ZS Avg.} case, with all
other models performing similarly to the Adam baseline. Counterintuitively,
trust region methods perform no better than SAM-style methods which do not
explicitly target representational similarity. This observation could be related
to recent insight into the smoothness side effects of training with SAM
\citep{wen2023sharpness-not-only-minimize}. We additionally raise that neither
metric in \Cref{tab:analysis_total} shows a similar trend to our empirical
findings---comparisons here do not strictly reflect similar performance
variation on specific tasks. Despite empirical improvement, recent work
questions if sharpness meaningfully correlates with generalization
\citep{juneja2023linear-connectivity-reveals,
pmlr-v202-andriushchenko23a-a-modern-look}. Extending TRAM should further
evaluate this relationship and investigate how trust region measurement could
inform better predictors of generalization capability.

\section{Conclusion}

We present TRAM: \textbf{T}rust \textbf{R}egion \textbf{A}ware
\textbf{M}inimization. TRAM optimizes for smoothness in both parameter and
function spaces to improve domain generalization during fine-tuning. TRAM
inherits the capability of SAM to optimize towards flatter minima and integrates
trust region awareness to ensure low local curvature between output
representations. We evaluate TRAM on out-of-distribution scenarios, where the
model must generalize to new distributions unseen during training. In this
setup, TRAM proves more effective than SAM-style optimization or trust region
methods. Our analysis identifies how TRAM bucks the anticorrelated trend for
major distribution shifts, learns a flatter out-of-domain loss surface, and
improves representation similarity for data unseen during fine-tuning.

\section{Acknowledgments}

TS gratefully acknowledges the support of the UK Engineering and Physical
Sciences Research Council (grant EP/W002876/1). This work has been made possible
in part by a gift from the Chan Zuckerberg Initiative Foundation to establish
the Kempner Institute for the Study of Natural and Artificial Intelligence.

\bibliography{main, anthology}
\bibliographystyle{iclr2024_conference}

\appendix

\section{Data Splits}\label{app:data}

\subsection*{Vision Datasets}

For vision modality experiments, we evaluate cross-dataset transfer from
ImageNet \citep{5206848-imagenet} to CIFAR-100
\citep{krizhevsky2009learning-cifar}, Stanford Cars
\citep{krause-2013-stanford-cars}, and Oxford Flowers
\citep{nilsback-2008-flowers}. We source all datasets from HuggingFace\footnote{
\texttt{huggingface.co/datasets/cifar100}\\
\quad \texttt{huggingface.co/datasets/Multimodal-Fatima/StanfordCars\_train}\\
\quad \texttt{huggingface.co/datasets/nelorth/oxford-flowers}} using the default
training/testing partitions.

\subsection*{Language Datasets}

We evaluate the M2D2 dataset \citep{reid-etal-2022-m2d2} for cross-domain
language modeling. M2D2 contains two groups: 11 domains from the S2ORC corpus of
ArXiv listings \citep{lo-etal-2020-s2orc} and an archive of Wikipedia articles.
We train a language model on each split's largest domain and evaluate zero-shot
generalization to ten domains unseen during fine-tuning. Evaluation uses
token-level perplexity across each domain. \Cref{tab:data:m2d2} details the
partition sizes (in tokens) for each domain in M2D2.

Zero-shot cross-lingual transfer is evaluated using MultiNLI and XNLI for
entailment classification. In this task, a model predicts an entailment label
(neutral, entailment, contradiction) between sentence pairs. We use only English
language MultiNLI \citep{williams-etal-2018-broad} for training data and
evaluate the trained model on the 14 unseen natural languages in XNLI
\citep{conneau-etal-2018-xnli} during test time. These datasets are balanced in
label classes and we report accuracy per language in our results. A complete
breakdown of partition sizes is shown in \Cref{tab:data:xnli}.

\begin{table}[t]
\centering
\caption{Data splits for M2D2 \citep{reid-etal-2022-m2d2} across Wikipedia and S2ORC \citep{lo-etal-2020-s2orc}. For simplicity, we do not consider the fine-grained subdomains in each domain. All data sourced from Huggingface (\texttt{huggingface.co/datasets/machelreid/m2d2})}
\label{tab:data:m2d2}
\resizebox{\textwidth}{!}{
\begin{tabular}{@{}cccccccc@{}}
\toprule
Split                  & Domain                          & Abbrev.     & Size (Tokens) & Training Domain & Train Tokens & Validation Tokens & Test Tokens \\ \midrule
\multirow{11}{*}{Wiki} & Culture and the arts            & {\sc Cult.} & 289M          &                 & ---          & ---               & 34.33M      \\
                        & General reference           & {\sc Gen.}    & 60M  &            & ---  & --- & 2.38M  \\
                        & Health and fitness          & {\sc Health.} & 116M &            & ---  & --- & 6.83M  \\
                        & History and events          & {\sc Hist.}   & 226M &            & ---  & --- & 11.65M \\
                        & Human activities            & {\sc Human.}  & 343M &            & ---  & --- & 12.41M \\
                        & Mathematics and logic       & {\sc Math.}   & 52M  &            & ---  & --- & 1.65M  \\
                       & Natural and physical sciences   & {\sc Nat.}  & 189M          &                 & ---          & ---               & 13.45M      \\
                        & Philosophy and thinking     & {\sc Phil.}   & 165M &            & ---  & --- & 2.32M  \\
                        & Religion and belief systems & {\sc Rel}     & 64M  &            & ---  & --- & 5.44M  \\
                       & Society and social sciences     & {\sc Soc.}  & 397M          & \checkmark      & 380M         & 11.8M                & 11.74M      \\
                       & Technology and applied sciences & {\sc Tech.} & 297M          &                 & ---          & ---               & 11.78M      \\ \midrule
\multirow{11}{*}{S2ORC} & Art                         & {\sc Art}     & 98M  &            & ---  & --- & 1.06M  \\
                        & Astrophysics                & {\sc Astro}   & 728M &            & ---  & --- & 1.14M  \\
                        & Condensed matter            & {\sc CondM.}  & 688M &            & ---  & --- & 1.17M  \\
                        & Computer science            & {\sc CS}      & 1.1B &            & ---  & --- & 1.17M  \\
                        & Economics                   & {\sc Econ.}   & 11M  &            & ---  & --- & 1.16M  \\
                        & Mathematics                 & {\sc Math}    & 1.4B & \checkmark & 1.1B & 1.46M  & 1.40M   \\
                        & Nonlinear sciences          & {\sc NLin.}   & 134M &            & ---  & --- & 1.28M  \\
                        & Philosophy                  & {\sc Phil.}   & 156M &            & ---  & --- & 1.06M  \\
                        & Physics                     & {\sc Phys.}   & 737M &            & ---  & --- & 1.12M  \\
                        & Quantitative biology        & {\sc QBio}    & 336M &            & ---  & --- & 1.08M  \\
                        & Statistics                  & {\sc Stat}    & 450M &            & ---  & --- & 1.19M  \\ \bottomrule
\end{tabular}%
}

\end{table}

\begin{table}[t]
\centering
\caption{Data splits for XNLI \citep{conneau-etal-2018-xnli}. The Training data in English is sourced from the MultiNLI dataset \citep{williams-etal-2018-broad} with translations provided for XNLI. Model selection during training uses only the English validation data. Validation data for other languages is used to measure $\epsilon$-sharness in our analysis. We omit data splits not used in this work. All data sourced from HuggingFace (\texttt{huggingface.co/datasets/xnli}).}
\footnotesize
\begin{tabular}{@{}ccccc@{}}
\toprule
XNLI                          &  Abbrev. & Train Sentences & Validation Sentences & Test Sentences \\ \midrule
English                       & {\sc En} & 393K            & 2.5K                 & 5K             \\
Arabic                        & {\sc Ar} & ---             & 2.5K                  & 5K             \\
Bulgarian                     & {\sc Bg} & ---             & 2.5K                  & 5K             \\
German                        & {\sc De} & ---             & 2.5K                  & 5K             \\
Greek                         & {\sc El} & ---             & 2.5K                  & 5K             \\
Spanish                       & {\sc Es} & ---             & 2.5K                  & 5K             \\
French                        & {\sc Fr} & ---             & 2.5K                  & 5K             \\
Hindi                         & {\sc Hi} & ---             & 2.5K                  & 5K             \\
Russian                       & {\sc Ru} & ---             & 2.5K                  & 5K             \\
Swahili                       & {\sc Sw} & ---             & 2.5K                  & 5K             \\
Thai                          & {\sc Th} & ---             & 2.5K                  & 5K             \\
Turkish                       & {\sc Tr} & ---             & 2.5K                  & 5K             \\
Urdu                          & {\sc Ur} & ---             & 2.5K                  & 5K             \\
Vietnamese                    & {\sc Vi} & ---             & 2.5K                  & 5K             \\
Chinese (Simplified) & {\sc Zh} & ---             & ---                  & 5K             \\ \bottomrule
\end{tabular}%
\label{tab:data:xnli}
\end{table}

\section{Additional Experimental Details}\label{app:experiments}


\subsection{Model training}
\label{app:experiments:design:training}

We fine-tune each pre-trained model without any freezing or additional
task-specific parameters where possible. We also do not explore fine-tuning with
low-rank approximations or adapters i.e., `full fine-tuning'. This setup
isolates the contribution of the optimization algorithm over additional capacity
in the model. For image classification and cross-lingual entailment
classification, we follow fine-tuning norms and only introduce a new
dataset-specific `head' to predict dataset-specific logits. For language tasks,
we fine-tune each pre-trained model for 50,000 steps using an initial learning
rate of $2\times 10^{-5}$, a polynomial decay schedule, and 10,000 step learning
rate warmup. We use Adam \citep{kingma2017adam}, with a decay factor setting
$\left(\beta_{1},\beta_{2}\right)=\left(0.9,0.99\right)$, as the base optimizer
for each SAM-style and TR method unless mentioned otherwise. When using
validation loss for model selection, we use only the validation partition of the
training domain to reflect a stricter evaluation setup without access to
additional domains during training. All models are trained $1\times$A100 80GB
GPU for under 72 hours except for GPT2-XL experiments in \Cref{app:lm-xl}.

\subsection{Baselines}
\label{app:experiments:design:baselines}

We compare to a naive SGD baseline for vision experiments following
\citet{pmlr-v162-kim22f-FisherSAM}. Our naive baseline for language experiments
is Adam \citep{kingma2017adam} without any augmentation setting decay factors as
$\left(\beta_{1},\beta_{2}\right)=\left(0.9,0.99\right)$. All algorithms listed
below use Adam as the inner optimizer for the final update (e.g.,
\Cref{alg:tram} Step 6).

For sharpness-aware methods: we compare to SAM ($\rho= 0.05$,
\citealp{foret2021sharpnessaware-SAM}), Adaptive SAM (ASAM, $\rho= 0.5$,
\citealp{pmlr-v139-kwon21b-ASAM}) and Fisher SAM (FSAM, $\gamma=0.1,~\eta=0.1$,
\citealp{pmlr-v162-kim22f-FisherSAM}).

For trust region methods: we compare to Trust Region Policy Optimization (TRPO,
\citealp{pmlr-v37-schulman15-trust-region-policy}), R3F ($\sigma=0.1$,
\citealp{aghajanyan2021better}), and MESA \citep{du2022sharpnessaware-for-free}.
MESA is a variant of TRPO regularizing output representation divergence between
current $\theta_{t}$ and the exponential moving average of previous
$\theta_{<t}$ with decay factor 0.999. For trust-region methods, we add the
regularizer directly to the task-specific loss function with a weighting
coefficient of $\lambda=0.1$ (in \Cref{eq:trust_region_1}). 

\subsection{Cross-dataset transfer for image classification}
\label{app:experiments:design:viz}

We implement the same cross-dataset adaptation setup as
\citet{pmlr-v162-kim22f-FisherSAM} as a `sanity check' directly comparing TRAM
to prior methods in the same setting. This setup is not strictly similar to the
`out-of-distribution' scenario we report for language tasks---this experiment
verifies that TRAM is performant on standard benchmarks and valuably evaluates
TRAM in the vision modality. The objective is to adapt ViT-base
\citep{dosovitskiy2021an-vit} from ImageNet pre-training
\citep{5206848-imagenet} to additional image classification tasks. We evaluate
dataset adaptation to CIFAR-100 \citep{krizhevsky2009learning-cifar}, Oxford
Flowers \citep{nilsback-2008-flowers} and Stanford Cars
\citep{krause-2013-stanford-cars} datasets. Our hypothesis is that TRAM can
improve applying information from ImageNet to additional datasets with different
labels and input data. 

We match the experimental setting of \citet{pmlr-v162-kim22f-FisherSAM}:
fine-tuning ViT-base-16 for 200 epochs with a base optimizer of SGD, an initial
learning rate of $5\times 10^{-4}$, and a cosine learning rate decay with no
warmup or restarts. We do not use early stopping to match prior work and use the
final model regardless of validation loss. We report the average Top-1 accuracy
over 5 runs, $\pm$ the 95\% confidence interval, in \Cref{tab:viz} for direct
comparison to \citet[Table 3]{pmlr-v162-kim22f-FisherSAM}. 

\subsection{Cross-domain language modeling} 
\label{app:experiments:design:lm}

We consider zero-shot cross-domain language modeling using the M2D2 Corpus
\citep{reid-etal-2022-m2d2}. Our hypothesis is that TRAM can better apply
language modeling information from large text corpora to improve out-of-domain
perplexity when fine-tuning to a specific domain.  For S2ORC, we train on the
``Math'' domain ({\sc Math}, 1.4B tokens) and for Wikipedia, we train on the
``Society and social sciences'' domain ({\sc Soc.}, 379M tokens). We use the
112M parameter GPT-2 base model \citep{radford2019language-gpt2} with a batch
size of 16 blocks of 1024 tokens following the setup of prior work
\citep{reid-etal-2022-m2d2, chronopoulou-etal-2022-efficient,
chronopoulou-etal-2023-adaptersoup}. We evaluate generalization via perplexity
for each test domain. We also evaluate a zero-shot baseline (i.e., GPT-2 before
fine-tuning) to contrast with the same model before domain-specific adaptation.
To reduce computation, we train one model with one random seed per algorithm. 

\subsection{Zero-shot cross-lingual transfer}
\label{app:experiments:design:xlt}

We test zero-shot cross-lingual transfer by fine-tuning a multilingual model on
an English task and then evaluating the model in other languages.  We
hypothesize that TRAM can improve task transfer across languages by improving
the usage of information from multilingual pre-training during monolingual
fine-tuning. A poorer model may `forget' other languages during the adaptation
process. We evaluate transfer from English to additional languages by predicting
labels for the XNLI test set after training the model for NLI only in English.
We use the 250M XLM-Roberta Base multilingual pre-trained model
\citep{conneau-etal-2020-unsupervised} with a classification head trained from
scratch. This model uses a batch size of 32 examples using only English
validation loss for model selection. Each reported result is averaged across 20
runs of varying random seeds to control for variation in loss surface.  

\subsection{Training algorithms}\label{app:tram_fisher}

\begin{algorithm}[t]
  \caption{Trust Region Aware Minimization}
  \label{alg:tram}
    \begin{footnotesize}
    \begin{algorithmic}
    \setcounter{ALC@unique}{0}
   \STATE {\bfseries Input:} Training set $S=\{(x_i,y_i)\}$, loss function $\ell$,
    learning rate $\alpha$, model parameters $\theta$,
   noise standard deviation $\sigma$ \algorithmiccomment{if noise-estimated
   trust region}.
   \FOR{$t=1,2,\dots$} 
   \STATE (1) Sample batch of $B=\lbrace\left(x_{i},~y_{i}\right)\rbrace_{i=0}^{|B|}$ data from $S$.
      \STATE (2) Compute the predictive distribution,
      $p_{f}\left(\cdot|x_{B},\theta_{t}\right)$, and gradient of the batch loss
      $\nabla L_B(\theta)$.
      \STATE (3) Compute trust region distance $d$ as:\\
      \ \ \ \ \ \ ${d}_{\theta}$ using $p_{f}\left(\cdot|x_{B},\theta_{t-1}\right)$ (\Cref{eq:trust_region_hist}) or\\
      \ \ \ \ \ \ 
        ${d}_{x}$ using $p_{f}\left(\cdot|x_{B}+z,\theta_{t}\right),~z\sim N\left(0,~\sigma^{2}\right)$ (\Cref{eq:trust_region_noise}).
      \STATE (4) Compute $\epsilon_{TRAM}^{\ast}$: \\
       \ \ \ \ \ \ $\epsilon^{\ast}_{\rm TRAM} =
\nicefrac{d~\theta^{2}\nabla L_{S}\left(\theta_{t}\right)}{||\theta \nabla L_{S}\left(\theta_{t}\right)||_{2}}$
      \STATE (5) Ascent step perturbing $\theta$ to $\theta +
      \epsilon_{TRAM}^{\ast}$.
      \STATE (6) Compute gradient at $\theta + \epsilon_{TRAM}^{\ast}$ as \Cref{eq:tram_final_grad}: \\
      \ \ \ \ \ \ $\nabla L_{\rm TRAM}\left(\theta\right) = \frac{\partial
      L_{S}}{\partial\theta}\biggr\rvert_{\theta=\theta+\epsilon^{\ast}_{\rm
      TRAM}}$
      \STATE (7) Gradient descent update: $\theta \leftarrow \theta - \alpha
      \nabla L_{\rm TRAM}(\theta)$.
   \ENDFOR
\end{algorithmic}
\end{footnotesize}
\end{algorithm}

\begin{algorithm}[t]
  \caption{Trust Region Aware Minimization with Fisher Information Matrix (TRAM-Fisher)}
  \label{alg:tramfisher}
    \begin{footnotesize}
    \begin{algorithmic}
    \setcounter{ALC@unique}{0}
   \STATE {\bfseries Input:} Training set $S=\{(x_i,y_i)\}$, loss function $\ell$,
    learning rate $\alpha$, model parameters $\theta$,
   noise standard deviation $\sigma$ 
   \FOR{$t=1,2,\dots$} \STATE 1) Sample batch of
   $B=\lbrace\left(x_{i},~y_{i}\right)\rbrace_{i=0}^{|B|}$ data from $S$.
      \STATE 2) Compute the predictive distribution,
      $p_{f}\left(\cdot|x_{B},\theta_{t}\right)$, and gradient of the batch loss
      $\nabla L_B(\theta)$.
      \STATE 3) Sample input noise $z\sim N\left(0,~\sigma^{2}I_{|\theta|}\right)$.
      \STATE 4) Approximate the Fisher Information Matrix at $x+z$: \\
        \ \ \ \ \ \  $\hat{F}\left(x+z;~\theta\right) = {\rm Diag}\left(\frac{1}{|B|} \sum_{i\in B} \left(\log p_{f}\left(y_{i}|x_{i}+z_{i}, \theta\right)\right)\right)^{2}$
      \STATE 5) Compute $\epsilon_{\rm TRAM-F}^{\ast}$: \\
      \ \ \ \ \ \ $\epsilon_{\rm \rm TRAM-F}^{\ast}= \frac{\hat{F}\left(x+z;~\theta\right)^{-1}\nabla L_{S}}{\sqrt{\nabla L_{S}\hat{F}\left(x+z;~\theta\right)^{-1}\nabla L_{S}}}$.
      \STATE 6) Ascent step perturbing $\theta$ to $\theta +
      \epsilon_{TRAM-F}^{\ast}$.
      \STATE 7) Compute gradient at $\theta + \epsilon_{TRAM-F}^{\ast}$ as \Cref{eq:tram_final_grad}: \\
      \ \ \ \ \ \ $\nabla L_{\rm TRAM-F}\left(\theta\right) = \frac{\partial
      L_{S}}{\partial\theta}\biggr\rvert_{\theta=\theta+\epsilon^{\ast}_{\rm
      TRAM}}$
      \STATE 8) Gradient descent update: $\theta \leftarrow \theta - \alpha
      \nabla L_{\rm TRAM-F}(\theta)$.
   \ENDFOR
\end{algorithmic}
\end{footnotesize}
\vspace{-0.1em}
\end{algorithm}

The training algorithm for TRAM is outlined in \Cref{alg:tram} using different
metrics for trust region estimation,~$d$, outlined in \Cref{sec:tram}.
\Cref{alg:tramfisher} details the TRAM-Fisher algorithm. Practically, this
modifies \Cref{alg:tram} in removing one forward pass to estimate the trust
region distance and instead approximate the Fisher Information Matrix of the
trust region neighborhood in representation space.

\subsection{Measuring sharpness}\label{app:sharpness_def} 

We follow \citet{keskar2017on} in evaluating model $\epsilon$-sharpness as
\Cref{eq:keskarsharpness} where $\ell$ is the loss function, $x \in
\mathbb{R}^{n}$ are $n$ model parameters, $A\in \mathbb{R}^{n\times~p}$ is a
matrix restricting the $\epsilon$-sharpness to a subspace of $p$ parameters
($A^{+}$ is the pseudo-inverse of $A$) and $\mathcal{C}_{\varepsilon}$ is
defined as \Cref{eq:sharpness_c} denoting a ``box'' region around the solution
over which loss is maximized.

\begin{align}
    \phi_{x,f}\left(\epsilon, A\right) &:= \frac{\max_{y\in \mathcal{C}_{\epsilon}}~\ell\left(x+Ay\right)-\ell\left(x\right)}{1+\ell\left(x\right)}\times 100
    \label{eq:keskarsharpness} \\
    \mathcal{C}_{\epsilon} &= \lbrace z \in \mathbb{R}^{p}: -\epsilon{\left(|{(A^{+}x)}_{i}|+1\right)} \leq z_{i} \leq \epsilon\left(|{(A^{+}x)}_{i}|+1\right) \forall~ i \in \left[ p \right] \rbrace \label{eq:sharpness_c}
\end{align}

For our measurement of $\epsilon$-sharpness, we set $A$ to the identity matrix
$I_{n\times n}$ to measure over the complete model. We measure
$\epsilon$-sharpness over the validation set of XNLI in all languages comparing
between original loss $\ell\left(x\right)$ and maximized loss $\max_{y\in
\mathcal{C}_{\epsilon}}~\ell\left(x+Ay\right)$. We follow the
$\epsilon$-sharpness setup of \citet{juneja2023linear-connectivity-reveals}
using an SGD optimizer, learning rate of $8\times 10^{-5}$, a 32 example batch
size, accumulation over 4 steps and $\epsilon$ of $1\times 10^{-5}$.

\subsection{Measuring representation similarity}\label{app:cka_def}

We follow \citet{pmlr-v97-kornblith19a} and \citet{conneau-etal-2020-emerging}
in evaluating cross-lingual similarity using Centered Kernel Alignment (CKA). At
a language level, CKA computes a similarity score between matrix $X$ and $Y$
where $X,Y\in \mathbb{R}^{n\times~d}$ are dense matrices of $n$ outputs of
$d$-dimensional representations from each model. We compute linear CKA
similarity as \Cref{eq:cka} using the Frobenius norm. For our cross-lingual
transfer experiments, we use the base model output for each example (i.e., the
representation before the classification head) to evaluate similarity.

\begin{equation}
    {\rm CKA}\left(X, Y \right)=\frac{{\lVert Y^{T}X\rVert}^{2}_{F}}{{\lVert X^{T}X \rVert}_{F}{\lVert Y^{T}Y\rVert}_{F}}
    \label{eq:cka}
\end{equation}

\section{Additional results}

\subsection{Domain correlations for S2ORC}\label{app:domain_correlation}

\begin{table}[ht]
\centering
\caption{Pearson correlation between training domains and zero-shot domains for M2D2. We report how the change in training domain correlates with changes in zero-shot perplexity to analyze how different domains improve or worsen during fine-tuning. All domains are correlated with {\sc Soc.} for the Wikipedia split. {\sc Art} and {\sc Phil.} domains are anti-correlated with {\sc Math} training domain for S2ORC indicating a major distribution shift.}
\label{tab:m2d2_domain_correlation}
\resizebox{\textwidth}{!}{%
\begin{tabular}{@{}ccc|cccc@{}} \toprule
Wiki Domain & $\rho$ to {\sc Soc.} & $p<0.01?$ & S2ORC Domain & $\rho$ to {\sc Math} & $p<0.01?$ & {\sc STEM}? \\ \midrule
{\sc Cult.} & 0.982 & \checkmark & {\sc Art} & -0.861 & \checkmark &  \\
{\sc Gen.} & 0.983 & \checkmark & {\sc Astro} & 0.812 & \checkmark & \checkmark \\
{\sc Health.} & 0.970 & \checkmark & {\sc CondM.} & 0.999 & \checkmark & \checkmark \\
{\sc Hist.} & 0.998 & \checkmark & {\sc CS} & 0.996 & \checkmark & \checkmark \\
{\sc Human.} & 0.980 & \checkmark & {\sc Econ.} & 0.997 & \checkmark & \checkmark \\
{\sc Math.} & 0.976 & \checkmark & {\sc NLin.} & 1.000 & \checkmark & \checkmark \\
{\sc Nat.} & 0.982 & \checkmark & {\sc Phil.} & -0.825 & \checkmark &  \\
{\sc Phil.} & 0.985 & \checkmark & {\sc Phys.} & 0.991 & \checkmark & \checkmark \\
{\sc Rel} & 0.994 & \checkmark & {\sc QBio} & 0.932 & \checkmark & \checkmark \\
{\sc Tech.} & 0.983 & \checkmark & {\sc Stat} & 0.998 & \checkmark & \checkmark \\ \midrule
{\sc ZS Avg.} & 0.990 & \checkmark & {\sc ZS Avg.} & 0.968 & \checkmark &  \\ \midrule
 &  &  & {\sc STEM Avg} & 0.998 & \checkmark &  \\ \bottomrule
\end{tabular}
}
\end{table}

\Cref{tab:m2d2_domain_correlation} details the correlation between zero-shot and
training domain perplexity across methods. We omit the combination approaches
(e.g., ASAM+R3F) due to poor performance. For Wikipedia, all domains are
correlated with the training domain indicating that the domain-specific fine
tuning on {\sc Soc.} domain has a net positive improvement on all zero-shot
domains. This trend is not consistent for S2ORC where we observe that {\sc Art}
and {\sc Phil.} domains are anti-correlated with the {\sc Math} training domain.
Improvement to {\sc Math} perplexity worsens the performance on these domains
across all methods. As discussed in \Cref{results:analysis:easy_and_hard}, TRAM
reports perplexity below this trend to perform better than expected for a
negatively correlated trend. For comparison, we contrast the correlations
between positively correlated domains (grouped as an average entitled {\sc
STEM}) and anticorrelated domains in \Cref{fig:distant-domains}.

\subsection{Training GPT2-XL with TRAM}\label{app:lm-xl}

\begin{table}[t]
\centering
\caption{M2D2 perplexity across training algorithms for GPT2-XL. We fine-tune on
the {\sc Math} domain M2D2 S2ORC split and evaluate in-domain and out-of-domain
perplexity. We evaluate TRAM, competitive comparisons and a GPT2-XL zero-shot
baseline. We omit algorithms demonstrating poorer results in smaller scale
experiments to limit computation demands. As in \Cref{tab:m2d2}, TRAM performs
strongly compared to all comparisons.  We report the average zero-shot
perplexity ({\sc ZS Avg.}) as the summary metric to judge domain transfer
capability (lower is better).  
Worst perplexity (excluding zero-shot) is \textcolor{minimum}{red}, best is
\textcolor{high}{green}. }
\label{tab:m2d2:s2orc:xl}
\resizebox{\textwidth}{!}{%
\begin{tabular}{@{}lc|cccccccccc|c@{}} 
\toprule
S2ORC & {\sc Math} & {\sc Art} & {\sc Phil.} & {\sc Astro} & {\sc CondM.} & {\sc CS} & {\sc Econ.} & {\sc NLin.} & {\sc Phys.} & {\sc QBio} & {\sc Stat} & {\sc ZS Avg.} $\downarrow$ \\ \midrule
GPT2-XL & 16.9 & 22.8 & 21.2 & 19.8 & 19.0 & 17.3 & 18.5 & 17.8 & 20.7 & 19.8 & 14.9 & 19.2 \\ \midrule
Adam & 8.7 & \colmin{30.4} & \colmin{28.2} & \colmin{24.0} & \colmin{14.9} & \colmin{15.4} & 15.4 & 11.4 & \colmin{21.4} & 22.1 & 12.6 & 19.6 \\
SAM & 8.7 & 29.3 & 28.0 & 22.6 & 14.6 & 15.1 & 15.1 & 11.2 & 20.5 & 21.4 & 12.3 & 19.0 \\
ASAM & 7.9 & 28.0 & 26.1 & 21.8 & 13.4 & 14.1 & 14.1 & 10.4 & 19.4 & 20.3 & 11.4 & 17.9 \\
FSAM & \colmax{7.8} & 26.7 & 25.0 & 21.1 & \colmax{13.1} & \colmax{13.7} & \colmax{13.7} & \colmax{10.2} &  \colmax{18.8} & 19.6 & \colmax{11.2} & \colmax{17.3} \\ \midrule
TRPO & 8.9 & 27.9 & 26.4 & 23.0 & \colmin{14.9} & 15.3 & 15.3 & 11.5 & 20.8 & 21.3 & 12.5 & 18.9 \\
R3F & 8.9 & 27.9 & 26.4 & 23.0 & \colmin{14.9} & 15.3 & 15.3 & 11.5 & 20.8 & 21.3 & 12.5 & 18.9 \\
MESA & \colmin{9.1} & 28.7 & 26.7 & 23.7 & 14.8 & 15.0 & \colmin{16.3} & \colmin{13.1} & 20.7 & \colmin{23.2} & \colmin{12.8} & 19.5 \\ \midrule
TRAM-${\theta_{t-1}}$ & 8.3 & \colmax{25.2} & \colmax{23.8} & \colmax{20.1} & 13.7 & 14.0 & 14.2 & 10.7 & 18.9 & \colmax{19.4} & 11.5 & \colmax{17.2} \\ \bottomrule
TRAM-${x}$ & 8.3 & 25.3 & \colmax{23.8} & 20.2 & 13.8 & 14.1 & 14.2 & 10.8 & 19.0 & 19.5 & 11.6 & \colmax{17.2} \\
\end{tabular}
}
\end{table}

\citet{bahri-etal-2022-sharpness} report that training with SAM is effective
over all sizes of T5 \citep{JMLR:v21:20-074}. We verify if this improvement
trend extends to TRAM by training a GPT2-XL model (1.5B parameters) on the same
language modeling task for 100,000 steps. The setup is the same as described in
\Cref{app:experiments} but we use 4 A100 GPUs for training each with a batch
size per device of 4 blocks $\times$ 1024 tokens. Perplexity for S2ORC domains
is shown in \Cref{tab:m2d2:s2orc:xl} where we observe similar trends to the 112M
parameter GPT2 model. We choose not to run these larger experiments on methods
with poor performance in \Cref{tab:m2d2} (e.g., combined approaches,
TRAM-Fisher) to limit computation demands. Zero-shot GPT2-XL is a stronger
baseline here which some methods struggle to improve upon despite improvement in
the training domain. TRAM-${\theta_{t-1}}$ and TRAM-${x}$ perform similarly
reporting the lowest perplexity in four domains. The most competitive adjacent
algorithm is FSAM reporting the lowest perplexity in seven domains. The
difference between FSAM and either TRAM algorithm is not significant here, as we
observed for smaller models in \Cref{tab:m2d2:s2orc}.

\subsection{Results from combining optimization algorithms}
\label{app:combo}
\begin{table}[ht]
  \caption{M2D2 perplexity (lower is better) on Wikipedia (upper) \& S2ORC
  (lower) splits. TRAM-${\theta_{t-1}}$ significantly improves over prior work
  ($p<0.01$ Kolmogorov-Smirnov test). Results are grouped as: (i) optimizers;
  (ii) trust region methods; (iii) combined SAM optimizers and trust region
  methods; and (iv) TRAM variants. The leftmost column is the training domain
  and we evaluate zero-shot perplexity on ten domains unseen during fine-tuning
  (full details in \Cref{app:data}). {\sc ZS Avg.} is the macro-average of all
  zero-shot domains.}
  \begin{subtable}[v]{\textwidth}
  \resizebox{\textwidth}{!}{%
    \begin{tabular}{@{}lc|cccccccccc|c@{}} \toprule
    \textbf{Wiki} & {\sc Soc.} & {\sc Cult.} & {\sc Gen.} & {\sc Health.} & {\sc Hist.} & {\sc Human.} & {\sc Math.} & {\sc Nat.} & {\sc Phil.} & {\sc Rel} & {\sc Tech.} & {\sc ZS Avg.} $\downarrow$ \\ \midrule
    GPT-2 & 27.2 & 27.7 & 27.8 & 24.5 & 29.2 & 28.8 & 28.6 & 29.4 & 27.8 & 27.7 & 28.7 & 28.0 \\  \midrule
    Adam & 24.8 & 26.3 & 26.4 & 23.6 & 27.2 & 27.0 & 27.4 & 27.6 & 26.3 & 25.8 & 27.4 & 26.5 \\
    SAM & 24.5 & 25.9 & 26.0 & 23.1 & 26.9 & 26.6 & 26.6 & 27.2 & 25.8 & 25.5 & 27.0 & 26.1 \\
    ASAM & 24.8 & 25.4 & 25.6 & 22.5 & 27.1 & 26.4 & 26.3 & 26.7 & 25.5 & 25.5 & \colmin{28.1} & 25.9 \\
    FSAM & 21.7 & 23.0 & 23.3 & 20.6 & 23.9 & 23.7 & 23.8 & 24.0 & 23.1 & 22.8 & 24.0 & 23.2 \\  \midrule
    TRPO & 21.8 & 23.0 & 23.3 & 20.7 & 24.0 & 23.7 & 23.8 & 24.0 & 23.1 & 22.8 & 24.1 & 23.3 \\
    R3F & 21.8 & 23.0 & 23.3 & 20.7 & 24.0 & 23.7 & 23.8 & 24.0 & 23.1 & 22.8 & 24.1 & 23.3 \\
    MESA & 23.1 & 24.0 & 24.3 & 21.5 & 25.4 & 24.9 & 24.8 & 25.2 & 24.1 & 24.0 & 25.1 & 24.3 \\  \midrule
    ASAM+TRPO & \colmin{25.6} & \colmin{26.8} & \colmin{26.9} & \colmin{24.0} & \colmin{28.0} & \colmin{27.6} & \colmin{27.6} & \colmin{28.2} & \colmin{26.8} & \colmin{26.5} & 27.9 & \colmin{27.0} \\
    ASAM+R3F & 25.0 & 26.0 & 26.2 & 23.2 & 27.4 & 26.9 & 26.8 & 27.4 & 26.1 & 25.9 & 27.1 & 26.3 \\
    ASAM+MESA & 25.3 & 26.3 & 26.5 & 23.5 & 27.7 & 27.2 & 27.1 & 27.7 & 26.3 & 26.1 & 27.4 & 26.6 \\  \midrule
    TRAM-${\theta_{t-1}}$ & \colmax{20.9} & \colmax{22.4} & \colmax{22.7} & \colmax{20.1} & \colmax{23.1} & \colmax{22.9} & \colmax{23.2} & \colmax{23.3} & \colmax{22.4} & \colmax{22.0} & \colmax{23.4} & \colmax{22.5} \\
    TRAM-${\theta_{0}}$ & 21.9 & 23.1 & 23.4 & 20.7 & 23.9 & 23.3 & 23.9 & 23.8 & 23.1 & 22.7 & 23.9 & 23.2 \\
    TRAM-${x}$ & 21.9 & 23.1 & 23.4 & 20.7 & 24.0 & 23.3 & 23.9 & 23.9 & 23.2 & 22.7 & 23.9 & 23.2 \\
    TRAM-Fisher & 22.5 & 23.7 & 24.0 & 21.3 & 24.6 & 24.0 & 24.7 & 24.6 & 23.8 & 23.3 & 24.6 & 23.9 \\ \bottomrule
    \end{tabular}
    } 
    \label{tab:m2d2:appfull:wiki}
  \end{subtable}
  
  \begin{subtable}[v]{\textwidth}
  \centering
    \resizebox{\textwidth}{!}{%
    \begin{tabular}{@{}lc|cccccccccc|c@{}}
    \toprule
  \textbf{S2ORC} &
    {\sc Math} &
    {\sc Art} &
    {\sc Astro} &
    {\sc CondM.} &
    {\sc CS} &
    {\sc Econ.} &
    {\sc NLin.} &
    {\sc Phil.} &
    {\sc Phys.} &
    {\sc QBio} &
    {\sc Stat} &
    {\sc ZS Avg.} $\downarrow$ \\ \midrule
  GPT-2   & 27.6 & 35.8    & 32.4    & 30.9 & 27.9 & 29.5 & 27.6 & 33.7    & 33.5 & 30.9 & 23.4 & 30.6 \\ \midrule
  Adam    & 11.4 & 44.2    & 33.9    & 20.1 & 21.2 & 21.0 & 14.7 & 41.9    & 29.5 & 30.8 & 16.9 & 27.4 \\
  SAM     & 10.5 & 45.3    & 33.2    & 18.7 & 20.3 & 20.0 & 13.7 & 42.4    & 28.3 & 30.2 & 16.1 & 26.8 \\
  ASAM    & 10.3 & 45.6    & 33.2    & 18.5 & 20.1 & 19.8 & 13.5 & 42.6    & 28.2 & 30.2 & 15.9 & 26.8 \\
  FSAM    & 10.4 & 45.6    & 33.3    & 18.5 & 20.2 & 19.9 & 13.5 & 42.7    & 28.3 & 30.2 & 15.9 & 26.8 \\ \midrule
  TRPO    & 10.4 & 46.0    & 33.4    & 18.6 & 20.3 & 20.0 & 13.6 & 42.9    & 28.4 & 30.4 & 16.0 & 26.9 \\
  R3F     & 10.4 & 46.0    & 33.4    & 18.6 & 20.2 & 20.0 & 13.6 & 42.9    & 28.4 & 30.4 & 16.0 & 26.9 \\
  MESA    & 11.9 & \colmax{44.1} & 34.1    & 20.8 & 21.7 & 21.6 & 15.3 & \colmax{41.7} & 30.0 & 31.0 & 17.4 & 27.8 \\ \midrule
  ASAM+TRPO &
    \colmin{13.7} &
    46.6 & \colmin{36.9} & \colmin{23.6} & \colmin{23.8} & \colmin{23.8} & \colmin{17.4} & \colmin{43.8} & \colmin{33.1} & \colmin{33.5} &
    \colmin{19.2} & \colmin{30.2} \\
  ASAM+R3F  & 13.5 & 46.2    & 36.5    & 23.3 & 23.6 & 23.5 & 17.2 & 43.4    & 32.7 & 33.2 & 19.0 & 29.9 \\
  ASAM+MESA     & 13.4 & 45.9    & 36.3    & 23.1 & 23.4 & 23.3 & 17.0 & 43.2    & 32.5 & 33.0 & 18.9 & 29.7 \\ \midrule
  TRAM-${\theta_{t-1}}$ & \colmax{9.6} & \colmin{46.8} & 32.5 & \colmax{17.2} & \colmax{19.2} & \colmax{18.9} & \colmax{12.6} & 43.3 & \colmax{27.0} & \colmax{29.6} & \colmax{15.0} & \colmax{26.2} \\
  TRAM-${\theta_{0}}$ & 10.4 & 44.8    & 33.0    & 18.6 & 20.1 & 19.9 & 13.6 & 42.0    & 28.2 & 30.0 & 15.9 & 26.6 \\
  TRAM-${x}$    & 10.4 & 44.9    & 33.0    & 18.6 & 20.1 & 19.9 & 13.6 & 42.0    & 28.1 & 30.0 & 15.9 & 26.6 \\
  TRAM-Fisher  & 10.5 & 46.1    & \colmax{32.4} & 18.7 & 20.3 & 20.0 & 13.6 & 43.0    & 28.2 & 30.3 & 16.0 & 26.9 \\ \bottomrule
  \end{tabular}
    } 
    \label{tab:m2d2:appfull:s2orc}
   \end{subtable}
   \label{tab:m2d2:appfull}
\end{table}

\begin{table}[t]
\centering
\caption{XNLI accuracy (higher is better) for training language ({\sc En}) and
14 zero-shot target languages summarised by {\sc ZS Avg.} (key in
\Cref{app:data}). All TRAM variants significantly outperform other methods
($p<0.01$ Wilcoxon test). Results are grouped as: (i) optimizers; (ii) trust
region methods; (iii) combined SAM optimizers and trust region regularization; and
(iv) TRAM variants. We report the mean across 20 seeds with standard deviation
in \Cref{tab:xnli_stddev}.} 

\resizebox{\textwidth}{!}{%
\begin{tabular}{@{}lc|cccccccccccccc|c@{}}
\toprule
 &
  {\sc En} &
  {\sc Ar} &
  {\sc Bg} &
  {\sc De} &
  {\sc El} &
  {\sc Es} &
  {\sc Fr} &
  {\sc Hi} &
  {\sc Ru} &
  {\sc Sw} &
  {\sc Th} &
  {\sc Tr} &
  {\sc Ur} &
  {\sc Vi} &
  {\sc Zh} &
  {\sc ZS Avg} $\uparrow$ \\ \midrule
Adam &
  \colmin{83.9} &
  \colmin{71.2} &
  \colmin{77.1} &
  \colmin{75.7} &
  \colmin{75.2} &
  \colmin{78.3} &
  77.6 &
  69.6 &
  \colmin{74.9} &
  64.6 &
  \colmin{71.2} &
  \colmin{72.2} &
  65.8 &
  74.1 &
  \colmin{73.1} &
  \colmin{72.9} \\
SAM                   & 84.8 & 72.1 & 78.1 & 76.7 & 75.7 & 79.0 & 77.9       & 69.8       & 75.7 & 65.2       & 71.8 & 73.1 & 66.8       & 75.1       & 74.2 & 73.7 \\
ASAM                  & 85.0 & 72.0 & 78.4 & 76.9 & 76.1 & 79.5 & 78.5       & 70.4       & 76.1 & 65.2       & 72.5 & 73.4 & 66.9       & 75.5       & 74.2 & 74.0 \\
FSAM                  & 84.7 & 72.2 & 78.1 & 76.9 & 76.0 & 79.3 & 78.4       & 70.0       & 76.1 & 65.1       & 72.2 & 73.0 & 66.8       & 75.3       & 74.2 & 73.8 \\ \midrule
TRPO                  & 84.9 & 71.3 & 77.7 & 76.2 & 75.3 & 78.6 & \colmin{77.3} & \colmin{69.2} & 75.2 & 64.4       & 71.6 & 72.4 & \colmin{65.3} & \colmin{73.8} & 73.3 & 73.0 \\
R3F                   & 85.5 & 72.7 & 78.9 & 77.5 & 76.8 & 79.9 & 79.2       & 70.7       & 76.8 & 66.2       & 72.9 & 73.9 & 66.6       & 75.8       & 74.6 & 74.5 \\
MESA                  & 84.9 & 71.9 & 77.9 & 76.7 & 75.7 & 78.8 & 77.8       & 69.6       & 75.8 & \colmin{64.1} & 72.1 & 72.4 & 65.7       & 74.4       & 73.9 & 73.3 \\ \midrule
ASAM+TRPO             & 85.0 & 72.4 & 78.5 & 77.2 & 76.4 & 79.7 & 78.9       & 70.4       & 76.4 & 65.3       & 72.4 & 73.2 & 66.8       & 75.7       & 74.6 & 74.1 \\
ASAM+R3F              & 85.1 & 72.1 & 78.3 & 76.9 & 75.9 & 79.3 & 78.4       & 70.3       & 76.0 & 65.1       & 72.4 & 73.3 & 66.3       & 75.1       & 74.3 & 73.8 \\
ASAM+MESA             & 84.7 & 71.7 & 77.8 & 76.3 & 75.7 & 78.8 & 77.9       & 69.5       & 75.4 & \colmin{64.1} & 71.6 & 72.7 & 65.6       & 74.3       & 73.4 & 73.2 \\ \midrule
TRAM-${\theta_{t-1}}$ & \colmax{86.2} & 73.1 & 79.5 & 78.2 & 77.0 & 80.2 & \colmax{79.7} & \colmax{71.5} & \colmax{77.5} & \colmax{66.4} & 73.3 & 74.2 & 67.5 & \colmax{76.7} & 75.8 & 75.0 \\
TRAM-${\theta_{0}}$ & 85.6 & 72.9 & 79.3 & 77.8 & 77.4 & 80.2 & 79.6       & 71.2       & 77.1 & 65.9       & 73.3 & 74.2 & 67.5       & \colmax{76.7} & 75.8 & 74.9 \\
TRAM-${x}$ & \colmax{86.2} & \colmax{73.5} & \colmax{79.8} & \colmax{78.3} & \colmax{77.5} & \colmax{80.9} & 79.6 & 71.4 & \colmax{77.5} & 66.0 & \colmax{73.8} & \colmax{74.3} & \colmax{67.6} & \colmax{76.7} & \colmax{75.9} & \colmax{75.2} \\
TRAM-Fisher           & 84.3 & 73.1 & 78.7 & 77.1 & 76.2 & 79.5 & 78.4       & 71.4       & 76.6 & 65.7       & 73.2 & 73.6 & 67.5       & 75.5       & 75.5 & 74.4 \\ \bottomrule
\end{tabular}%
}
\label{tab:xnli:appfull}
\end{table}

Given that TRAM builds on integrating SAM-style optimization with trust-region
regularization, we additionally compare to a naive combination of each of these
methods. We replace the standard loss function in ASAM with the loss function
adding trust region regularization. 

Our full results featuring these systems are shown in \Cref{tab:m2d2:appfull}
for language modeling and \Cref{tab:xnli:appfull} for zero-shot cross-lingual
transfer. Across both tasks, naive combination approaches are some of the
weakest approaches. When we directly combine ASAM with each trust region
regularizer (TRPO, R3F, MESA), we find that the naive combination approaches
perform worse than Adam alone, even with extensive hyperparameter tuning. We
conjecture that the constituent methods fail to compound constructively because
the trust region regularizer does not interact with (or respect) the $\rho$-ball
neighborhood of ASAM. Therefore, each component may contribute to cross-feature
interference, with a disadvantageous net effect on training. TRAM instead offers
to combine strategies with complementary features without interference. 

\subsection{Run variation in cross-lingual transfer}

For XNLI experiments, we report the mean over 20 runs varying random seed in
\Cref{tab:xnli} and \Cref{tab:xnli:distance}. We report the respective standard
deviation values for each reported mean in \Cref{tab:xnli_stddev}.

\begin{table}[t]
\centering
\caption{Standard deviation of accuracy for the XNLI dataset across 20 training runs with varying random seed. Results are split into groups for: (i) optimizers, (ii) trust region methods, (iii) combined methods, (iv) TRAM variants and (v) TRAM using $d_{x}$ with varying metrics for computing divergence. This accompanies \Cref{tab:xnli} and \Cref{tab:xnli:distance} which report average values across seeds.}
\resizebox{\textwidth}{!}{%
\begin{tabular}{@{}lccccccccccccccc@{}}
\toprule
 & {\sc En} & {\sc Bg} & {\sc De} & {\sc El} & {\sc Ar} & {\sc Es} & {\sc Fr} & {\sc Hi} & {\sc Ru} & {\sc Sw} & {\sc Th} & {\sc Tr} & {\sc Ur} & {\sc Vi} & {\sc Zh} \\ \midrule
Adam                    & 0.34 & 0.42 & 0.65 & 0.47 & 0.50 & 0.39 & 0.51 & 0.51 & 0.51 & 0.65 & 0.40 & 0.41 & 0.43 & 0.51 & 0.55 \\
SAM                     & 0.24 & 0.31 & 0.35 & 0.34 & 0.31 & 0.32 & 0.49 & 0.33 & 0.50 & 0.36 & 0.40 & 0.35 & 0.44 & 0.39 & 0.39 \\
ASAM                    & 0.33 & 0.33 & 0.45 & 0.39 & 0.47 & 0.36 & 0.51 & 0.44 & 0.44 & 0.51 & 0.56 & 0.42 & 0.47 & 0.45 & 0.50 \\
FSAM                    & 0.35 & 0.31 & 0.51 & 0.56 & 0.35 & 0.37 & 0.47 & 0.41 & 0.45 & 0.53 & 0.37 & 0.39 & 0.44 & 0.35 & 0.38 \\ \midrule
TRPO                    & 0.24 & 0.35 & 0.37 & 0.34 & 0.30 & 0.39 & 0.34 & 0.46 & 0.40 & 0.38 & 0.36 & 0.34 & 0.53 & 0.38 & 0.34 \\
R3F                     & 0.34 & 0.40 & 0.44 & 0.38 & 0.35 & 0.35 & 0.43 & 0.42 & 0.46 & 0.41 & 0.41 & 0.35 & 0.43 & 0.39 & 0.47 \\
MESA                    & 0.34 & 0.34 & 0.44 & 0.24 & 0.40 & 0.52 & 0.37 & 0.67 & 0.43 & 0.26 & 0.40 & 0.45 & 0.59 & 0.43 & 0.34 \\ \midrule
ASAM+TRPO               & 0.30 & 0.28 & 0.36 & 0.29 & 0.26 & 0.28 & 0.35 & 0.34 & 0.44 & 0.36 & 0.39 & 0.38 & 0.34 & 0.32 & 0.32 \\
ASAM+R3F                & 0.32 & 0.45 & 0.45 & 0.40 & 0.46 & 0.36 & 0.49 & 0.53 & 0.50 & 0.52 & 0.40 & 0.49 & 0.60 & 0.45 & 0.49 \\
ASAM+MESA               & 0.34 & 0.31 & 0.30 & 0.44 & 0.42 & 0.39 & 0.38 & 0.51 & 0.58 & 0.46 & 0.44 & 0.31 & 0.51 & 0.50 & 0.46 \\ \midrule
TRAM-${\theta_{t-1}}$ & 0.40 & 0.31 & 0.40 & 0.30 & 0.36 & 0.31 & 0.43 & 0.50 & 0.53 & 0.48 & 0.43 & 0.34 & 0.36 & 0.49 & 0.42 \\
TRAM-${\theta_{0}}$   & 0.34 & 0.38 & 0.41 & 0.44 & 0.48 & 0.40 & 0.43 & 0.53 & 0.63 & 0.47 & 0.66 & 0.39 & 0.57 & 0.50 & 0.54 \\
TRAM-${x}$            & 0.31 & 0.29 & 0.45 & 0.44 & 0.37 & 0.33 & 0.38 & 0.37 & 0.48 & 0.39 & 0.44 & 0.32 & 0.43 & 0.39 & 0.42 \\
TRAM-Fisher             & 0.29 & 0.65 & 0.67 & 0.55 & 0.58 & 0.60 & 0.49 & 0.72 & 0.69 & 0.64 & 0.86 & 0.49 & 0.68 & 0.55 & 0.73 \\ \midrule
TRAM-${x}$ (MMD)      & 0.42 & 0.38 & 0.46 & 0.43 & 0.47 & 0.35 & 0.42 & 0.43 & 0.48 & 0.44 & 0.59 & 0.49 & 0.36 & 0.37 & 0.59 \\
TRAM-${x}$ ($L_{2}$)  & 0.30 & 0.27 & 0.27 & 0.26 & 0.24 & 0.28 & 0.27 & 0.21 & 0.29 & 0.26 & 0.24 & 0.28 & 0.21 & 0.22 & 0.22 \\ \bottomrule
\end{tabular}%
}
\label{tab:xnli_stddev}
\end{table}

\subsection{Choosing a distance metric}\label{app:xnli_distance_metric}

TRAM relies on KL divergence to estimate the trust region around the pre-trained
function (i.e., $p_{f}\left(\cdot|x+z, \theta\right)$ or $p_{f}\left(\cdot|x,
\theta_{t-1}\right)$). We propose TRAM with forward KL on the intuition that the
perturbed distribution (i.e., estimated point in the trust region) is the target
(i.e., true) output which the current outputs (i.e., estimate) should match. We
empirically verify this setup as the optimal arrangement (i.e., forward KL).
While reverse KL or symmetric KL report only marginally poorer results, we
report only forward KL for simplicity. We also consider alternative distance
metrics in \Cref{tab:xnli:distance}. We evaluate modifying the best-performing
model for XNLI with different distances to examine if the divergence for trust
region estimation is influential in performance. We evaluate maximum mean
discrepancy using an inverse multiquadratic kernel \citep[MMD;
][]{JMLR:v13:gretton12a-MMD}, or $L_{2}$ distance within $d_{x}$. Even using the
worst-performing metric, $L_{2}$ distance, TRAM is still competitive to methods
in \Cref{tab:data:xnli}. Characterizing the best trust region estimate for TRAM
is outside the scope of this work. Future work should explore the suitability of
different distances (e.g., Renyi divergence) to improve the estimation of the
trust region space. 

\begin{table}[t]
\centering
\caption{XNLI accuracy across varying the divergence metric estimating the trust region distance in TRAM. We compare to using maximum mean discrepancy (MMD) and $L_{2}$ distance. 
TRAM is generally robust to different estimates for the trust region between $p_{f}\left(\cdot|x, \theta\right)$ and $p_{f}\left(\cdot|x+z, \theta\right)$.
}
\resizebox{\textwidth}{!}{%
\begin{tabular}{lc|cccccccccccccc|c}
\toprule
 &
  {\sc En} &
  {\sc Bg} &
  {\sc De} &
  {\sc El} &
  {\sc Ar} &
  {\sc Es} &
  {\sc Fr} &
  {\sc Hi} &
  {\sc Ru} &
  {\sc Sw} &
  {\sc Th} &
  {\sc Tr} &
  {\sc Ur} &
  {\sc Vi} &
  {\sc Zh} &
  {\sc ZS Avg.} $\uparrow$ \\ \midrule
TRAM-${x}$ (KL) &
  \colmax{86.2} &
  \colmax{79.8} &
  \colmax{78.3} &
  \colmax{77.5} &
  \colmax{73.5} &
  \colmax{80.9} &
  \colmax{79.6} &
  \colmax{71.4} &
  \colmax{77.5} &
  \colmax{66.0} &
  73.8 &
  74.3 &
  \colmax{67.6} &
  \colmax{76.7} &
  \colmax{75.9} &
  \colmax{75.2} \\
TRAM-${x}$ (MMD) &
  86.0 &
  79.3 &
  78.1 &
  77.1 &
  73.2 &
  80.7 &
  \colmax{79.6} &
  \colmax{71.4} &
  77.3 &
  \colmax{66.0} &
  \colmax{74.0} &
  \colmax{74.4} &
  67.2 &
  76.3 &
  75.6 &
  75.0 \\
TRAM-${x}$ ($L_{2}$) &
  85.1 &
  78.7 &
  76.8 &
  76.2 &
  72.2 &
  79.4 &
  78.8 &
  70.4 &
  76.2 &
  65.5 &
  72.6 &
  73.5 &
  67.1 &
  75.8 &
  74.6 &
  74.1 \\ \bottomrule
\end{tabular}%
}
\vspace{-1em}
\label{tab:xnli:distance}
\end{table}

\end{document}